\PassOptionsToPackage{table}{xcolor}
\documentclass[sigconf]{acmart}

\usepackage{graphicx}
\usepackage{tabularx}
\usepackage{xcolor}
\usepackage{pifont}
\usepackage{booktabs}
\usepackage{multirow}
\usepackage[ruled,lined,linesnumbered]{algorithm2e}
\SetAlgoSkip{}
\SetAlgoInsideSkip{smallskip}
\usepackage{eurosym}
\usepackage{enumitem}
\microtypesetup{expansion=false}

\usepackage{array}

\usepackage{hyperref}
\hypersetup{
    colorlinks=true,
    urlcolor=blue
}

\usepackage{booktabs}
\usepackage{tabularx}
\usepackage{array}
\newcolumntype{Y}{>{\raggedright\arraybackslash}X}

\newcommand{\cmark}{{\color{green!70!black}\ding{51}}}
\newcommand{\xmark}{{\color{red}\ding{55}}}

\AtBeginDocument{%
  }

\setcopyright{none}
\copyrightyear{2026}
\acmYear{2026}
\acmDOI{}
\acmConference[Preprint]{Revised manuscript following peer review}{2026}{}
\acmISBN{}
\begin{document}

\title{AgentFAIR: A Multi-Agent Collaborative Framework for FAIRness Evaluation of Geospatial Datasets}

\author{Ming Chen}
\orcid{0000-0002-6120-6630}
\authornote{Both authors contributed equally to this work.}
\authornote{Corresponding author: Ming Chen (ming.chen6@unimelb.edu.au)}
\affiliation{%
  \institution{The University of Melbourne}
  \city{Melbourne}
  \state{Victoria}
  \country{Australia}
}

\author{Pranav Pai}
\authornotemark[1]
\affiliation{%
  \institution{The University of Melbourne}
  \city{Melbourne}
  \state{Victoria}
  \country{Australia}
}


\begin{abstract}
Geospatial datasets support critical applications ranging from urban
planning to climate modeling, yet consistently assessing their FAIR
maturity remains difficult. Existing evaluators differ in their rubrics
and evidence sources and may fail on JavaScript-rendered pages or
repository-specific identifier schemes. In a diagnostic study of 50
datasets from 10 repositories, the mean within-dataset standard
deviation of normalized scores across available tools is 15.0 points
on the 0--100 scale, with a maximum of 30.3 points. Because these
tools operationalize FAIR differently, we use their normalized scores
to characterize disagreement and failure modes rather than comparative
accuracy.

We present \textsc{AgentFAIR}, a multi-agent framework that combines
structured metadata extraction with 13 large language model evaluators,
one for each FAIR sub-principle. Each evaluator produces a 0--3 maturity
score, provenance-linked evidence, and recommendations; a critic checks
evidence sufficiency and cross-principle consistency and may trigger
targeted re-evaluation. Across the sample, mean Findability,
Accessibility, Interoperability, and Reusability scores are 79.7\%,
70.4\%, 45.3\%, and 72.0\%, respectively. Across five runs on a stratified 10-dataset subset, mean exact sub-principle agreement was 89\% (SD = 3 percentage points),
compared with 71\% (SD = 5 percentage points) without the critic.
In a preliminary study of 15 datasets, \textsc{AgentFAIR} matches the
majority expert label on 82\% of the 195 sub-principle judgments;
expert inter-rater agreement is Fleiss' $\kappa=0.71$. The mean LLM
API cost is USD \$0.054 per dataset, and the mean processing time is
1,054 seconds. The framework therefore provides evidence-linked,
auditable assessments at low model-inference cost; however, the
limited benchmark, incomplete component-level ablations, and
validation within a single model family limit conclusions about
accuracy and generalizability.
\end{abstract}



\maketitle

\section{Introduction}

The FAIR data principles---\textit{Findability}, \textit{Accessibility}, \textit{Interoperability}, and \textit{Reusability}---provide widely adopted guidance for making digital research outputs more reusable and machine-actionable \cite{wilkinson2016fair}.
In geospatial science, FAIRness (FAIR compliance; \emph{not} algorithmic fairness) is especially consequential: spatial layers, rasters, and spatio-temporal products are routinely integrated across agencies, jurisdictions, and scientific communities.
However, many geospatial datasets remain only partially FAIR, due to inconsistent metadata practices, heterogeneous formats (e.g., Shapefile, GeoTIFF, NetCDF), and incomplete spatial reference information \cite{thompson2023proposed,paprotny2023population}.
According to a European Commission cost-benefit analysis, non-FAIR practices are estimated to cost the EU research economy \EUR{10.2} Billion annually through duplicated effort and failed reuse \cite{pwc2018nonfair}.

Existing automated FAIR assessment tools such as F-UJI \cite{devaraju2020fujifair} and FAIR-Checker \cite{gaignard2023fair} provide important mechanistic checks (e.g., persistent identifiers, licenses, basic schema markup). However, deterministic approaches face limitations that are amplified for geospatial resources: (i) \textbf{Dynamic metadata exposure:} Repository landing pages increasingly rely on JavaScript rendering and API-driven content, which breaks traditional crawlers and causes partial or missing evidence collection; 
(ii) \textbf{Repository heterogeneity:} Identifier and metadata practices vary widely (DOI, Handle, internal URIs; different schema profiles), thus hard-coded assumptions (e.g., DataCite-centric resolution patterns) can lead to brittle failures; 
(iii) \textbf{Limited semantic judgment:} Rule-based checks cannot reliably assess qualitative and domain-specific aspects of metadata, such as whether a coordinate reference system is valid, whether access constraints are optional or mandatory, or whether terminology reflects community standards; 
(iv) \textbf{Low score comparability:} Recent analyses show that different tools operationalize FAIR differently, leading to substantial score variance and making stewardship decisions tool-dependent \cite{candela2024fair,krans2022fair}.

Furthermore, geospatial FAIRness analysis depends on domain conventions that general-purpose tools often treat as opaque strings.
Examples include ISO (International Organisation for Standardisation)~19115 metadata profiles \cite{iso19115-12014}, coordinate reference system identifiers (EPSG codes), and OGC discovery and access conventions (e.g., WMS, WFS, and STAC endpoints).
Because many geospatial vocabularies are not registered in common ontology registries, semantic validators may systematically underestimate \textit{Interoperability}, even when datasets follow community practice.

\begin{figure*}[t]
    \centering
    \includegraphics[width=0.9\textwidth]{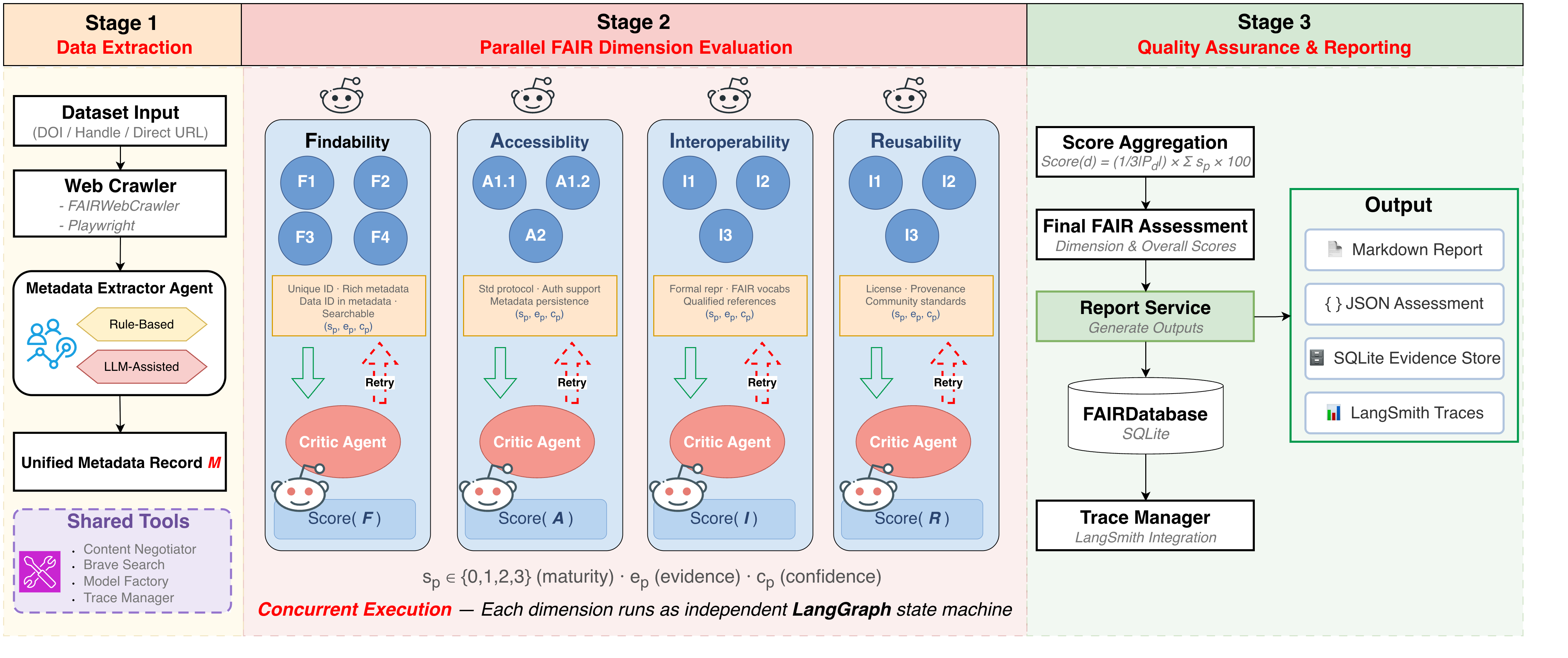}
    \Description{System diagram showing AgentFAIR's pipeline: URL input, orchestration, agent execution with critic loop, persistence of audit trails, and reporting.}
    \caption{The proposed \textsc{AgentFAIR} framework for multi-agent collaborative FAIRness evaluation.}
    \label{fig:Architecture}
\end{figure*}

To address these challenges, we propose \textsc{AgentFAIR}, a multi-agent collaborative framework that combines:
(i) deterministic extraction of machine-readable metadata and identifiers, and
(ii) sub-principle-specific LLM agents that perform evidence-grounded semantic evaluation. As shown in Figure~\ref{fig:Architecture}, the system orchestrates a pipeline that processes dataset landing pages, extracts metadata, evaluates each FAIR sub-principle with specialized agents, and applies critic-based quality control via running critic agents to ensure evidence-backed scoring and consistency across principles. When evidence is insufficient or internally inconsistent, the critic agent triggers a targeted re-evaluation (\textit{Retry} mechanism) with refined instructions, improving robustness while preserving traceability.
Overall, \textsc{AgentFAIR} evaluates all 13 FAIR sub-principles (F1--F4, A1.1--A2, I1--I3, R1.1--R1.3) on a 0--3 maturity scale and outputs a Markdown report, JSON assessment, SQLite evidence store, and local execution traces. Remote LangSmith tracing is optional and disabled by default.

We evaluate \textsc{AgentFAIR} on 50 geospatial datasets across 10 repositories, including Zenodo\footnote{\url{https://zenodo.org} (Open repository for research outputs with DOI assignment.)},
Dryad\footnote{\url{https://datadryad.org} (Curated repository for datasets supporting published scientific research.)},
Harvard Dataverse\footnote{\url{https://dataverse.harvard.edu} (General-purpose repository for sharing and citing research data.)},
PANGAEA\footnote{\url{https://www.pangaea.de} (Earth and environmental science repository for georeferenced datasets.)},
NOAA NCEI\footnote{\url{https://www.ncei.noaa.gov} (Global archive of climate, oceanic, and environmental data.)},
and domain-specific portals.
We compare against four widely used FAIR evaluators (F-UJI~\cite{devaraju2020fujifair},
FAIR-Checker~\cite{gaignard2023fair}, FAIRshake~\cite{clarke2019fairshake},
FAIR-enough~\cite{fair_enough}) and analyze
(i) overall FAIR maturity patterns,
(ii) cross-tool agreement and failure modes,
(iii) cost and runtime characteristics,
and (iv) consistency across repeated evaluations.

To the best of our knowledge, \textsc{AgentFAIR} is the first multi-agent LLM-based framework specifically designed for FAIR compliance evaluation. It integrates critic-informed feedback loops with domain-specific geospatial reasoning. The main contributions of this work are as follows:

\begin{itemize}
    \item \textbf{The first LLM-driven multi-agent FAIR evaluator.} We propose \textsc{AgentFAIR}, a collaborative multi-agent framework with critic-based quality control and explicit awareness of geospatial standards. To our knowledge, it is the first system to combine LLM reasoning with domain-specific FAIR assessment.
    \item \textbf{A transparent and fine-grained scoring protocol.} We operationalize a 0--3 maturity rubric for each FAIR sub-principle and explicitly map geospatial indicators to the assessment of \textit{Interoperability} and \textit{Reusability}.
    \item \textbf{A cross-repository empirical study.} We evaluate 50 geospatial datasets across 10 repositories. The sample reveals strong \textit{Findability} (79.7\%) and weak \textit{Interoperability} (45.3\%), but is not intended to establish long-tail generalization.
    \item \textbf{A diagnostic analysis of evaluator disagreement and feasibility.} We quantify cross-tool disagreement (mean per-dataset standard deviation 15.0 points; maximum 30.3), document repository-specific failure modes, and measure an average API cost of approximately USD \$0.054 per dataset. These comparisons do not treat heterogeneous tool scores as equivalent accuracy measures.
\end{itemize}

The rest of the paper is organized as follows. Section~2 reviews related work on FAIR assessment tools, geospatial data quality, and multi-agent LLM architectures. Section~3 formalizes the task definition and scoring protocol. Section~4 details the \textsc{AgentFAIR} framework design and implementation. Section~5 presents the evaluation and its limitations, and Section~6 concludes the paper. The software artifact, dataset list, prompt structure, and LangGraph orchestration details are provided in the Appendix. The source code is available at \url{https://doi.org/10.5281/zenodo.18529560} and \url{https://github.com/MingCHEN-Github/AgentFAIR}.

\section{Related Work}

\subsection{Automated FAIR Assessment Tools}
Automated FAIR evaluators typically operationalize the principles as predefined, machine-checkable indicators (e.g., presence of PIDs, protocol support, and machine-readable licensing) and aggregate the results into per-principle scores.
F-UJI implements the FAIRsFAIR assessment metrics and collects evidence from dataset landing pages and exposed metadata, complemented by queries to external registries \cite{devaraju2020fujifair}.
FAIR-Checker emphasizes semantic validation by leveraging knowledge graphs and SHACL/SPARQL constraints \cite{gaignard2023fair}.
FAIRshake provides community-curated rubrics tailored to different digital resource types \cite{clarke2019fairshake}, whereas FAIR-enough offers a broader suite of largely binary tests, which can be conservative when evidence is partial or indirectly expressed \cite{fair_enough}.
A recurring challenge is that tools differ in their operational definitions, evidence sources, and aggregation rules, leading to scores that are not directly comparable across evaluators \cite{candela2024fair}.

Table~\ref{tab:tool-comparison} summarizes representative capabilities of these systems relative to \textsc{AgentFAIR}.
For baselines, \emph{coverage} denotes whether a tool---or our mapping of its outputs---supports assessment across all 13 FAIR sub-principles used in this paper.

\begin{table}[t]
\centering
\caption{Comparison of FAIR assessment systems}
\label{tab:tool-comparison}
\resizebox{\columnwidth}{!}{%
\begin{tabular}{lcccccc}
\toprule
\textbf{System} & \textbf{LLM} & \textbf{Geo-aware} & \textbf{Explain} & \textbf{Trace} & \textbf{Cross-P} & \textbf{Coverage} \\
\midrule
F-UJI \cite{devaraju2020fujifair} & \xmark & \xmark & \cmark & \cmark & \xmark & \cmark \\
FAIRshake \cite{clarke2019fairshake} & \xmark & \xmark & \cmark & \cmark & \xmark & \xmark \\
FAIR-Checker \cite{gaignard2023fair} & \xmark & \xmark & \cmark & \cmark & \cmark & \xmark \\
FAIR-enough \cite{fair_enough} & \xmark & \xmark & \cmark & \cmark & \xmark & \cmark \\
\textbf{AgentFAIR} & \cmark & \cmark & \cmark & \cmark & \cmark & \cmark \\
\bottomrule
\end{tabular}%
}
\par\smallskip
{\footnotesize LLM=LLM-based reasoning in the evaluation loop; Geo-aware=explicit recognition of geospatial standards; Explain=evidence-backed scoring rationales; Trace=persisted audit trails; Cross-P=cross-principle consistency checks; Coverage=ability to support assessment across all 13 sub-principles used here.}
\end{table}

\subsection{FAIRness and Quality for Geospatial Data}
FAIRness assessment for geospatial datasets intersects with classic spatial data quality concerns, including spatial reference consistency, resolution/scale, temporal extent, and adherence to domain standards.
Metadata standards such as ISO~19115 \cite{iso19115-12014} enable rich descriptions of spatial resources, but repositories vary substantially in how completely these metadata elements are captured and exposed in machine-readable form.
As a result, automated checkers that rely on generic schemas, brittle pattern matching, or incomplete registry coverage may miss geospatial signals (e.g., CRS (Coordinate Reference System) identifiers or geospatial service endpoints), which can systematically depress \textit{Interoperability}---and, downstream, \textit{Reusability}---scores for spatial datasets.

\subsection{LLM Agent for Data Management and Evaluation}
Large language models (LLMs) have shown promise for extracting, normalizing, and enriching metadata from heterogeneous sources \cite{busch2025exploring,zhang2025data}.
Recent systems combine LLM reasoning with workflow orchestration to automate data cleaning, validation, and quality assessment \cite{li2024autodcworkflow,huang2024cocoon,narayan2024datallm}.
However, applying LLMs to FAIR evaluation introduces additional requirements---consistent rubrics, evidence-backed justifications, and auditable trails---and systematic multi-agent designs that explicitly address these requirements, especially under geospatial standards, remain limited.

\subsection{Multi-agent LLM Architectures}
Multi-agent frameworks decompose complex tasks into coordinated subtasks executed by specialized agents, often improving robustness through structured interaction and self-critique \cite{xi2024agentsurvey,wang2024llmagenteval}.
AutoGen \cite{wu2024autogen} introduced conversational multi-agent patterns enabling flexible agent interactions, while MetaGPT \cite{hong2024metagpt} demonstrated role-based collaboration for software engineering tasks.
Benchmarks such as AgentBench \cite{liu2024agentbench} and TheAgentCompany \cite{xu2024theagentcompany} evaluate agent capabilities on consequential real-world tasks, providing evaluation ideas that we adapt for FAIR compliance assessment.
For geospatial domains, foundation model applications \cite{mai2024geoai} highlight both the potential and the domain-specific challenges of applying LLMs to spatial information understanding.
Building on these architectural patterns, \textsc{AgentFAIR} combines critic-informed feedback loops with evidence-grounded prompting and explicit geospatial standard awareness to support transparent, auditable FAIR scoring.

\begin{table*}[t]

\centering
\footnotesize
\setlength{\tabcolsep}{2pt}        
\renewcommand{\arraystretch}{1.05} 
\begin{tabularx}{\textwidth}{@{}l l p{2.1cm} X X p{2.4cm}@{}}

\toprule
\textbf{Dim} & \textbf{ID} & \textbf{FAIR Principle} & \textbf{Evaluation Criteria (evidence)} & \textbf{Maturity Levels (0$\to$3)} & \textbf{Geo Indicators} \\
\midrule
\multicolumn{6}{@{}l@{}}{\textbf{\textit{Findability} (F)}}\\
\midrule
F & \textbf{F1} & Globally unique persistent identifier &
PID presence; HTTPS resolution; machine-readable embedding; content negotiation &
None $\to$ Unrecognized/fails $\to$ Resolves HTTPS $\to$ +Metadata+ContentNeg &
DOI, NASA CMR, STAC IDs \\
F & \textbf{F2} & Rich metadata description &
Core citation (creator, title, date, publisher) + descriptive (abstract, keywords, coverage) &
None $\to$ Minimal (title) $\to$ Core citation $\to$ Rich descriptive &
ISO 19115, spatial extent \\
F & \textbf{F3} & Metadata includes data identifier &
Data access links in metadata; multiple access methods; machine-readable distribution &
None $\to$ Present/broken $\to$ Resolvable $\to$ Multiple clear links &
OPeNDAP, WMS/WFS, STAC \\
F & \textbf{F4} & Registered in searchable resource &
Catalog registration (DataCite); Schema.org markup; API-verified indexing &
None $\to$ Meta tags only $\to$ Catalog OR structured $\to$ Both &
CKAN API, CSW, STAC \\
\midrule
\multicolumn{6}{@{}l@{}}{\textbf{\textit{Accessibility} (A)}}\\
\midrule
A & \textbf{A1.1} & Standardized retrieval protocol &
Protocol openness (HTTP/S, FTP); IETF/ISO standardization; programmatic access &
Proprietary $\to$ Restricted $\to$ Standard+limits $\to$ Open+free &
OGC, STAC API, THREDDS \\
A & \textbf{A1.2} & Authentication or authorization support &
Auth clarity (OAuth, API keys); automation feasibility; documentation quality &
None/broken $\to$ Complex $\to$ Standard auth $\to$ Open/clear &
Earthdata Login, ESA Hubs \\
A & \textbf{A2} & Metadata preserved without data &
Preservation via PIDs (DOI/ARK); repository certification (CoreTrustSeal); policy &
None $\to$ Basic repo $\to$ Institutional $\to$ Certified+policy &
PANGAEA, SDI catalogs \\
\midrule
\multicolumn{6}{@{}l@{}}{\textbf{\textit{Interoperability} (I)}}\\
\midrule
I & \textbf{I1} & Formal knowledge representation &
W3C/ISO standards (JSON-LD, RDF); recognized vocabulary usage; semantic richness &
None $\to$ Limited $\to$ Standard+vocab $\to$ Rich semantic &
GeoSPARQL, DCAT, Schema.org \\
I & \textbf{I2} & FAIR vocabularies used &
Vocabulary FAIRness: registry presence (LOV, BioPortal, BARTOC); documentation &
None $\to$ 1 registry $\to$ 2+ registries $\to$ 3+ with docs &
GCMD, CF conventions, SWEET \\
I & \textbf{I3} & Qualified references to other data &
Explicit relationships (derivedFrom, cites); scientific context; PIDs for refs &
None $\to$ Vague ``see also'' $\to$ Explicit typed $\to$ Machine-readable &
DataCite relations, PROV-O \\
\midrule
\multicolumn{6}{@{}l@{}}{\textbf{\textit{Reusability} (R)}}\\
\midrule
R & \textbf{R1.1} & Clear data usage license &
Standard license (SPDX/CC/ODbL); machine-readable embedding in metadata &
None $\to$ Free text $\to$ Standard ID $\to$ Machine-readable &
OGL, CC-BY, ODbL \\
R & \textbf{R1.2} & Detailed provenance &
Processing lineage (ISO/PROV-O); creator IDs (ORCID); methodology; versions &
None $\to$ Minimal $\to$ Clear derivation $\to$ Structured PROV-O &
ISO 19115 lineage, PROV-O \\
R & \textbf{R1.3} & Domain community standards &
Metadata standard compliance; controlled vocabularies; validated format specs &
None $\to$ Terminology $\to$ Recognizable std $\to$ Multiple formal &
ISO 19139, CF-NetCDF, STAC \\
\bottomrule
\end{tabularx}
\caption{Operationalization of the 13 FAIR sub-principles used in this paper: evidence criteria, 0--3 maturity levels, and representative geospatial indicators.}
\label{tab:rubric-mapping}
\end{table*}

\section{Task Definition and Scoring Protocol}

\subsection{FAIR Evaluation Task and Principle Set}
We formulate FAIR evaluation as an evidence-grounded auditing task \cite{wilkinson2016fair}.
Given a dataset landing-page URL $u$ (the human-facing entry point for a dataset) and optional repository context $m$ (e.g., repository type, known metadata schemas, and API conventions), \textsc{AgentFAIR} produces three structured outputs:
(i) per-sub-principle maturity scores $S=\{s_p\}_{p\in\mathcal{P}}$,
(ii) per-sub-principle evidence bundles $E=\{e_p\}_{p\in\mathcal{P}}$ that justify each score with provenance, and
(iii) actionable recommendations $R=\{r_p\}_{p\in\mathcal{P}}$ describing what is missing and how to improve compliance.

\textbf{Evidence model.}
For each sub-principle $p\in\mathcal{P}$, the evidence bundle $e_p$ is a structured record that includes (a) extracted metadata fields (e.g., PID, license, protocol), (b) supporting excerpts/snippets (e.g., HTML fragments or metadata blocks), and (c) provenance pointers (e.g., source URL and the extraction route such as landing page, embedded JSON-LD, downloadable metadata files, or repository API).
This design is critical for explainability and auditability: non-zero scores must cite observable, traceable evidence.

\textbf{Evaluated principles.}
We evaluate the 13 FAIR sub-principles from the FAIR Guiding Principles \cite{wilkinson2016fair}:
\begin{itemize}
    \item \textbf{\textit{Findability}:} F1 (globally unique persistent identifier), F2 (rich metadata), F3 (identifier in metadata), F4 (indexed in a searchable resource).
    \item \textbf{\textit{Accessibility}:} A1.1 (retrieval protocol), A1.2 (authentication/authorization), A2 (metadata persistence).
    \item \textbf{\textit{Interoperability}:} I1 (formal knowledge representation), I2 (FAIR vocabularies), I3 (qualified references).
    \item \textbf{\textit{Reusability}:} R1.1 (license), R1.2 (provenance), R1.3 (community standards).
\end{itemize}

\textbf{Role of repository context $m$.}
When available, $m$ provides lightweight repository-specific hints (e.g., typical locations of machine-readable metadata, common field names, or standardized API endpoints).
This improves recall of evidence without changing the rubric: \textsc{AgentFAIR} still assigns scores strictly according to the same maturity criteria, and all inferences must be backed by retrievable evidence recorded in $E$.

\subsection{Rubric-Based Maturity Scoring and Aggregation}
\textbf{Maturity rubric.}
Each sub-principle $p$ receives an ordinal maturity score $s_p\in\{0,1,2,3\}$:
\begin{itemize}
    \item \textbf{0 (non-compliant):} no supporting evidence is found, or evidence is contradictory/unverifiable from accessible resources.
    \item \textbf{1 (partial):} some relevant information exists but is incomplete, implicit, or primarily human-readable (weak machine actionability).
    \item \textbf{2 (substantial):} clear evidence exists in a machine-actionable form for the core requirement, but with notable gaps (e.g., missing qualifiers, incomplete standard alignment, or partial exposure across distributions).
    \item \textbf{3 (full):} strong, unambiguous evidence satisfies the requirement in a standard, machine-actionable way with sufficient detail for automated reuse and validation.
\end{itemize}
Table~\ref{tab:rubric-mapping} specifies the concrete indicators and decision rules used for each sub-principle. This explicit rubric is applied uniformly across repositories to reduce tool-specific interpretation variance \cite{wilkinson2019evaluating}.

\textbf{Dimension and overall aggregation.}
Let $\mathcal{D}=\{\text{F},\text{A},\text{I},\text{R}\}$ denote the four FAIR dimensions, and let $\mathcal{P}_d\subset\mathcal{P}$ be the proper subset of sub-principles within dimension $d$.
We report scores as percentages for comparability with existing evaluators:
\[
\mathrm{Score}(d)=\frac{1}{3|\mathcal{P}_d|}\sum_{p\in\mathcal{P}_d} s_p \times 100,
\qquad
\mathrm{Score}(\mathrm{FAIR})=\frac{1}{3|\mathcal{P}|}\sum_{p\in\mathcal{P}} s_p \times 100.
\]
Because $s_p$ is evidence-based, missing machine-readable signals directly translate into lower maturity and generate corresponding recommendations $r_p$.

\textbf{Geospatial operationalization.}
A common failure mode of generic FAIR checkers for spatial datasets is missing domain-specific signals that are essential for \textit{Interoperability} and \textit{Reusability}.
To mitigate this, \textsc{AgentFAIR} explicitly searches for and reasons over geospatial indicators when scoring I1--I3 and R1.3, including (non-exhaustively): coordinate reference system identifiers, spatial and temporal extent descriptors, ISO~19115-1 elements, and OGC-related service endpoints and encodings across different geospatial semantics. These indicators are not ``bonus points''; rather, they serve as admissible evidence under the rubric when they satisfy the corresponding sub-principle requirements (e.g., machine-interpretable representations for I1 and community standards for R1.3).

\textbf{Evaluation objectives supported by the protocol.}
The above task formulation and rubric are designed to support:
(i) \textbf{Consistency} (controlled variability under repeated runs despite LLM stochasticity),
(ii) \textbf{Explainability} (scores justified by explicit evidence and decision rules),
(iii) \textbf{Auditability} (persisted evidence and traces enabling post-hoc inspection), and
(iv) \textbf{Domain awareness} (recognition of geospatial standards and metadata constructs).

\section{Proposed AgentFAIR Framework}
\label{sec:framework}
Figure~\ref{fig:Architecture} summarizes \textsc{AgentFAIR} as a three-stage, evidence-first pipeline for evaluating all 13 FAIR sub-principles on the 0--3 maturity rubric defined in Section~3.
Given a dataset input (e.g., DOI, Handle, direct URL), the system executes:
\textbf{(Stage 1) Data extraction} $\rightarrow$
\textbf{(Stage 2) Parallel FAIR dimension evaluation} $\rightarrow$
\textbf{(Stage 3) Quality assurance and reporting}.

\textbf{Stage 1 (Data extraction).}
A Playwright-based crawler resolves redirects and renders JavaScript-heavy landing pages, then a hybrid extractor constructs a unified metadata record $M$ that merges structured fields (e.g., JSON-LD/Schema.org, HTML meta tags, DCAT-like fragments) with provenance-linked text snippets.
Agents must cite this record when assigning a non-zero score. The critic checks for missing evidence and conflicts with deterministic observations; this is a structural check on the returned record, not a claim that prompting alone guarantees factual correctness.

\textbf{Stage 2 (Parallel dimension evaluation).}
\textsc{AgentFAIR} decomposes evaluation into four concurrent LangGraph state machines, one per FAIR dimension (F/A/I/R), each internally evaluating its sub-principles in parallel.
For each sub-principle $p\in\mathcal{P}$, the corresponding agent outputs a tuple $(s_p, e_p, c_p)$ where
$s_p\in\{0,1,2,3\}$ is the maturity score,
$e_p$ is an evidence bundle (fields/snippets + provenance pointers),
and $c_p\in[0,1]$ is a confidence estimate used for quality control.

\textbf{Stage 3 (Quality assurance \& reporting).}
A critic-driven feedback loop checks evidence sufficiency and cross-sub-principle consistency before scores are aggregated into dimension and overall percentages (Section~3.2).
The report service then materializes auditable artifacts: a human-readable Markdown report, a machine-readable JSON assessment, a SQLite evidence store, and local execution traces for post-hoc inspection. LangSmith export is an explicit opt-in because prompts and evidence may contain sensitive content.

Algorithm~\ref{alg:agentfair} summarizes the control flow, including caching of deterministic signals to stabilize retries.
\begin{algorithm}[t]
\caption{\textsc{AgentFAIR} evaluation pipeline}\label{alg:agentfair}
\SetKwInOut{Input}{Input}
\SetKwInOut{Output}{Output}
\Input{Dataset URL $u$, repository context $m$}
\Output{Sub-principle scores $S$, evidence $E$, recommendations $R$}

$\mathit{html} \leftarrow \textsc{Crawl}(u)$\;
\tcc{Playwright for JS rendering}

$M \leftarrow \textsc{ExtractMetadata}(\mathit{html}, m)$\;
\tcc{Hybrid extraction \& enrichment}

$G \leftarrow \textsc{DeterministicChecks}(u,M)$\;
\tcc{Resolver, protocol, content-negotiation, and registry signals}

\ForEach{sub-principle $p \in \mathcal{P}$}{
    $(s_p, e_p, c_p) \leftarrow \textsc{Agent}_p(M,G)$\;
    \tcc{Evidence-backed evaluation}
    \If{$c_p < \theta_p$ \textbf{or} $\textsc{HardCheckFails}(p,s_p,e_p,G)$}{
        $(s_p, e_p, c_p) \leftarrow \textsc{CriticRepair}(p,M,G,s_p,e_p)$\;
        \tcc{Retry with feedback}
    }
}

$S \leftarrow \textsc{Aggregate}(\{s_p\})$\;
\tcc{Dimension and overall scores}

$R \leftarrow \textsc{GenerateRecommendations}(S, E)$\;
\Return{$S, E, R$}
\end{algorithm}













\subsection{LangGraph orchestration}
We use LangGraph \cite{LangGraph} to represent the multi-agent workflow as a directed state machine. Each FAIR dimension is expressed as a modular subgraph with nodes for crawling, extraction, sub-principle evaluation, critic review, and report synthesis. Conditional edges implement retry logic when confidence is low or evidence is missing (Appendix~\ref{app:langgraph}).
Checkpointing supports recovery, and trace integration enables inspection of agent decisions and intermediate state.

\subsection{Hybrid metadata extraction}
The Extractor stage combines:
(1) \textbf{rule-based parsing} of structured metadata (e.g., Schema.org/JSON-LD, DCAT-like fragments, HTML meta tags) and identifier patterns (DOI/Handle/ORCID); and
(2) \textbf{LLM-assisted enrichment} for normalization, deduplication, and resolving ambiguous natural-language fields (e.g., license statements and access conditions).
The output is a unified internal metadata record $M$ that retains both structured fields and provenance links to their sources (e.g., landing page text, embedded JSON-LD).

\subsection{Sub-principle agents}
\textsc{AgentFAIR} comprises 13 specialized agents, one per FAIR sub-principle.
Each agent receives the extracted metadata record $M$ and returns:
(i) a maturity score $s_p\in\{0,1,2,3\}$,
(ii) evidence $e_p$ consisting of specific supporting snippets or fields, and
(iii) a confidence score $c_p$ used for quality control.
Each system prompt separates six elements: (i) the sub-principle definition and scope; (ii) the 0--3 decision rubric; (iii) admissible deterministic and extracted evidence; (iv) geospatial indicators such as EPSG, ISO~19115, and OGC terms where relevant; (v) conservative instructions not to infer absent evidence; and (vi) a structured output schema requiring score, rationale, evidence, confidence, and recommendations. Retry prompts append the critic's identified contradiction or missing-evidence requirement while retaining the original rubric and cached deterministic observations.

\subsection{Critic agent and consistency checks}
The critic agent applies two forms of quality control:
\textbf{evidence sufficiency} (is the score supported by extracted evidence?) and
\textbf{cross-sub-principle consistency} (do scores contradict each other given the evidence?).
For example, high A1.1 (open protocol) paired with low A1.2 (authentication clarity) can indicate ambiguous access conditions; the critic triggers a targeted re-evaluation with refined instructions.
For Findability, seven deterministic checks include identifier presence, resolver success for maturity $\geq2$, content negotiation evidence for maturity 3, score-range validity, and evidence presence for non-zero scores. A retry can be triggered by a failed hard check or by confidence below 0.5 (0.4 for I1 in the reported configuration), subject to a retry budget. Evaluator and critic use the same GPT-4o-mini backend with different roles and prompts; therefore the critic does not remove shared-model bias. All retries and revisions are logged as part of the audit trail.

\subsection{Implementation notes}
\textsc{AgentFAIR} uses Playwright for JavaScript-rendered access and Extruct, lxml, and rdflib for structured metadata parsing. Geospatial standards and CRS identifiers are recognized from exposed metadata; the evaluated implementation does not perform GDAL-based inspection of downloaded data files.
SQLite persists evidence and local traces, and FastAPI exposes a REST API for local integration.
The primary experiment uses GPT-4o-mini with temperature 0.1. The non-zero setting was intended to permit limited lexical variation in rationales and evidence text; a temperature 0.0 versus 0.1 ablation was not conducted, so no performance benefit is claimed.

\subsection{Security boundary}
Dataset pages and metadata are untrusted input. The implementation restricts evaluation targets to public HTTP(S) destinations, uses hardened XML parsing, and requires structured outputs, but it has not been evaluated against prompt injection or poisoned metadata. In particular, text retrieved from a page can still influence an LLM evaluator. The released API is therefore intended for local, single-user operation; adversarial prompt-injection testing, redirect-by-redirect destination validation, authentication, rate limiting, and per-request credential isolation are required before public deployment.


\begin{table*}[t]
\centering
\scriptsize
\setlength{\tabcolsep}{1.1pt}
\renewcommand{\arraystretch}{1.0}
\caption{FAIR maturity matrix for 50 datasets across 13 sub-principles using the 0--3 rubric (representative run). Columns D1--D50 map to the dataset list in Appendix~\ref{app:datasets}.}
\label{tab:fair-maturity-matrix}
\resizebox{\textwidth}{!}{%
\begin{tabular}{@{}l*{50}{c}@{}}
\toprule
\textbf{Sub-principle} & \textbf{D1} & \textbf{D2} & \textbf{D3} & \textbf{D4} & \textbf{D5} & \textbf{D6} & \textbf{D7} & \textbf{D8} & \textbf{D9} & \textbf{D10} & \textbf{D11} & \textbf{D12} & \textbf{D13} & \textbf{D14} & \textbf{D15} & \textbf{D16} & \textbf{D17} & \textbf{D18} & \textbf{D19} & \textbf{D20} & \textbf{D21} & \textbf{D22} & \textbf{D23} & \textbf{D24} & \textbf{D25} & \textbf{D26} & \textbf{D27} & \textbf{D28} & \textbf{D29} & \textbf{D30} & \textbf{D31} & \textbf{D32} & \textbf{D33} & \textbf{D34} & \textbf{D35} & \textbf{D36} & \textbf{D37} & \textbf{D38} & \textbf{D39} & \textbf{D40} & \textbf{D41} & \textbf{D42} & \textbf{D43} & \textbf{D44} & \textbf{D45} & \textbf{D46} & \textbf{D47} & \textbf{D48} & \textbf{D49} & \textbf{D50} \\
\midrule
F1 & 1 & 0 & 3 & 2 & 3 & 3 & 3 & 3 & 3 & 1 & 3 & 3 & 3 & 1 & 1 & 1 & 3 & 3 & 3 & 3 & 1 & 3 & 3 & 0 & 1 & 3 & 3 & 3 & 3 & 1 & 3 & 3 & 1 & 3 & 1 & 2 & 3 & 2 & 3 & 3 & 2 & 3 & 1 & 1 & 3 & 3 & 1 & 1 & 1 & 1 \\
F2 & 3 & 2 & 3 & 3 & 3 & 3 & 3 & 3 & 3 & 3 & 3 & 3 & 3 & 3 & 3 & 3 & 3 & 3 & 3 & 3 & 3 & 3 & 3 & 2 & 3 & 3 & 3 & 3 & 3 & 3 & 3 & 3 & 3 & 3 & 3 & 3 & 3 & 3 & 3 & 3 & 3 & 3 & 3 & 3 & 3 & 2 & 3 & 3 & 3 & 3 \\
F3 & 3 & 0 & 3 & 3 & 3 & 3 & 2 & 0 & 2 & 1 & 1 & 3 & 3 & 1 & 1 & 0 & 2 & 3 & 3 & 3 & 1 & 2 & 2 & 0 & 2 & 3 & 2 & 3 & 3 & 1 & 3 & 3 & 1 & 3 & 1 & 1 & 3 & 3 & 1 & 1 & 1 & 2 & 2 & 1 & 3 & 0 & 0 & 0 & 1 & 2 \\
F4 & 3 & 3 & 3 & 3 & 3 & 3 & 2 & 3 & 3 & 3 & 2 & 3 & 3 & 3 & 2 & 3 & 3 & 2 & 3 & 3 & 2 & 3 & 3 & 1 & 3 & 2 & 2 & 3 & 3 & 2 & 2 & 2 & 3 & 3 & 2 & 3 & 3 & 3 & 3 & 2 & 3 & 3 & 3 & 3 & 2 & 2 & 3 & 2 & 2 & 3 \\
\midrule
A1.1 & 3 & 3 & 3 & 3 & 3 & 3 & 3 & 3 & 3 & 1 & 3 & 3 & 3 & 3 & 1 & 3 & 3 & 3 & 3 & 3 & 3 & 3 & 1 & 3 & 3 & 3 & 3 & 3 & 3 & 3 & 3 & 2 & 2 & 1 & 1 & 3 & 3 & 3 & 3 & 3 & 3 & 3 & 3 & 3 & 2 & 3 & 1 & 1 & 1 & 3 \\
A1.2 & 3 & 3 & 3 & 3 & 3 & 3 & 3 & 3 & 3 & 1 & 3 & 3 & 3 & 3 & 1 & 3 & 2 & 3 & 3 & 3 & 3 & 3 & 1 & 3 & 1 & 3 & 3 & 3 & 3 & 3 & 3 & 3 & 3 & 3 & 3 & 3 & 3 & 3 & 3 & 3 & 3 & 3 & 1 & 1 & 1 & 3 & 3 & 1 & 1 & 1 \\
A2 & 1 & 0 & 1 & 1 & 1 & 1 & 1 & 0 & 1 & 1 & 2 & 1 & 1 & 1 & 1 & 0 & 2 & 2 & 2 & 1 & 1 & 1 & 2 & 0 & 2 & 1 & 1 & 1 & 1 & 0 & 2 & 2 & 2 & 2 & 1 & 1 & 1 & 2 & 1 & 1 & 1 & 1 & 1 & 2 & 2 & 0 & 0 & 1 & 2 & 1 \\
\midrule
I1 & 2 & 0 & 2 & 3 & 2 & 2 & 2 & 0 & 2 & 2 & 2 & 2 & 3 & 2 & 2 & 0 & 3 & 2 & 2 & 2 & 2 & 2 & 3 & 0 & 3 & 2 & 2 & 2 & 2 & 2 & 2 & 2 & 2 & 2 & 2 & 2 & 2 & 2 & 2 & 2 & 2 & 2 & 3 & 3 & 3 & 0 & 0 & 0 & 2 & 2 \\
I2 & 1 & 0 & 1 & 1 & 1 & 1 & 1 & 0 & 1 & 1 & 1 & 1 & 1 & 1 & 1 & 0 & 2 & 1 & 1 & 1 & 1 & 1 & 2 & 0 & 2 & 1 & 1 & 1 & 1 & 1 & 1 & 1 & 1 & 1 & 2 & 2 & 2 & 2 & 1 & 1 & 1 & 1 & 1 & 1 & 1 & 0 & 0 & 0 & 1 & 1 \\
I3 & 2 & 0 & 1 & 2 & 2 & 2 & 2 & 0 & 1 & 1 & 1 & 2 & 2 & 2 & 1 & 0 & 1 & 1 & 1 & 1 & 1 & 1 & 1 & 0 & 1 & 1 & 2 & 2 & 2 & 1 & 1 & 1 & 1 & 2 & 2 & 2 & 2 & 2 & 1 & 1 & 1 & 1 & 1 & 1 & 1 & 0 & 0 & 1 & 1 & 1 \\
\midrule
R1.1 & 3 & 2 & 1 & 3 & 3 & 3 & 3 & 0 & 3 & 3 & 3 & 3 & 3 & 3 & 3 & 0 & 1 & 0 & 0 & 1 & 3 & 3 & 3 & 2 & 3 & 3 & 3 & 3 & 3 & 3 & 0 & 0 & 0 & 3 & 3 & 3 & 3 & 3 & 3 & 3 & 3 & 3 & 1 & 3 & 3 & 0 & 0 & 3 & 3 & 3 \\
R1.2 & 2 & 2 & 2 & 2 & 2 & 2 & 2 & 2 & 2 & 2 & 2 & 2 & 2 & 2 & 2 & 2 & 2 & 2 & 2 & 2 & 3 & 2 & 2 & 2 & 2 & 2 & 2 & 2 & 2 & 2 & 2 & 2 & 2 & 2 & 2 & 2 & 2 & 2 & 2 & 2 & 2 & 2 & 2 & 2 & 2 & 3 & 2 & 2 & 2 & 2 \\
R1.3 & 2 & 2 & 2 & 2 & 2 & 2 & 3 & 2 & 2 & 2 & 2 & 3 & 2 & 2 & 2 & 2 & 2 & 2 & 3 & 2 & 2 & 2 & 2 & 2 & 2 & 3 & 3 & 3 & 3 & 3 & 2 & 2 & 2 & 3 & 2 & 2 & 2 & 2 & 2 & 2 & 2 & 2 & 2 & 2 & 3 & 1 & 2 & 2 & 2 & 2 \\
\bottomrule
\end{tabular}%
}
\end{table*}

\begin{table}[t]
\centering
\caption{Case study FAIR dimension scores (\%) from the representative run in Table~\ref{tab:fair-maturity-matrix}.}
\label{tab:case-studies}
\resizebox{\columnwidth}{!}{%
\begin{tabular}{lccccr}
\toprule
\textbf{Dataset} & \textbf{F} & \textbf{A} & \textbf{I} & \textbf{R} & \textbf{Overall} \\
\midrule
AURIN OSM POIs (D2) & 41.7 & 66.7 & 0.0 & 66.7 & 43.6 \\
NASA GDIS (D3) & 100.0 & 77.8 & 44.4 & 55.6 & 71.8 \\
Zenodo Crater Lake (D1) & 83.3 & 77.8 & 55.6 & 77.8 & 74.4 \\
\bottomrule
\end{tabular}%
}
\end{table}

\section{Evaluation}
Experiments were conducted on an Apple M2 Pro system (32\,GB RAM, macOS~15).
The framework evaluated all 13 FAIR sub-principles using the 0--3 maturity rubric in Section~3.
LLM calls used \texttt{gpt-4o-mini} (temperature 0.1).
Playwright handled browser-based crawling (including JavaScript-rendered pages); SQLite stored evidence traces and audit logs.

\subsection{Datasets}
We evaluated 50 geospatial datasets spanning 10 repositories to capture heterogeneous metadata practices and access patterns.
Our benchmark prioritizes \emph{cross-repository diversity} over single-repository depth. It samples heterogeneous repository and domain conditions but is too small to establish long-tail generalization.
\textbf{Repository distribution:} Zenodo (10), Dryad (8), PANGAEA (7), Harvard Dataverse (6), NOAA NCEI (6), ScienceBase (4), Figshare (4), AURIN (2), EarthData (2), NASA SEDAC (1).
\textbf{Domains:} Biodiversity/Ecology (12), Marine/Geoscience (10), Climate/Hydrology (8), Urban/Land Classification (6), Demographics (4), Disaster/Emissions (3), Other (7).
Datasets are referenced as D1--D50 in Tables~\ref{tab:fair-maturity-matrix} and~\ref{tab:tool-score-matrix}; the full mapping appears in Appendix~\ref{app:datasets}.

\subsection{Baselines and score normalization}
We compare against four public FAIR evaluators: F-UJI, FAIR-Checker, FAIRshake, and FAIR-enough.
These tools expose different rubrics and output schemas (binary tests, rubric scores, or per-dimension scores), so we normalize each tool's \emph{overall} dataset score to a 0--100 scale.
For tools with multiple tests, the normalized score is the fraction of tests passed; for tools with per-dimension scores, we use the mean across dimensions.
For \textsc{AgentFAIR}, the overall score is computed from its 13 sub-principle maturities (0--3), scaled to 0--100.
This normalization supports only dataset-level disagreement and failure-mode analysis; it does \emph{not} make the underlying metrics equivalent and is not used as evidence that one evaluator is more accurate.

Baseline tools also differ in metadata acquisition strategy.
F-UJI relies primarily on API/identifier lookups, while \textsc{AgentFAIR} uses browser-based crawling via Playwright.
Browser-based extraction better reflects end-user access to JavaScript-rendered metadata, but introduces different failure modes (e.g., navigation timeouts on complex pages vs.\ API endpoint unavailability).

\textbf{Run selection and table consistency.}
Because \textsc{AgentFAIR} is LLM-based, it exhibits run-to-run variability (Section~\ref{sec:consistency}). We report one representative run in Table~\ref{tab:fair-maturity-matrix}; the \textsc{AgentFAIR} row in Table~\ref{tab:tool-score-matrix} is the direct aggregation of those same 13 scores for each dataset.

\subsection{Results: FAIR maturity patterns}

Table~\ref{tab:fair-maturity-matrix} reports maturity scores for each dataset (columns) across the 13 FAIR sub-principles (rows).
Aggregating these sub-principle scores, \textit{Findability} performs best on average (79.7\%), followed by \textit{Reusability} (72.0\%) and \textit{Accessibility} (70.4\%); \textit{Interoperability} is consistently weakest (45.3\%).

\textbf{Sub-principle bottlenecks.}
F2 (rich metadata) is the strongest sub-principle (mean 2.94/3; 47/50 datasets achieve full maturity), reflecting widespread use of descriptive landing pages and repository-generated metadata.
In contrast, A2 (metadata should remain accessible even when the data are no longer available) remains weak (mean 1.14/3; 0/50 at full maturity), reflecting limited explicit preservation policies, certification, and machine-actionable tombstone guarantees.
The sub-principles I2 and I3 from \textit{Interoperability} are the dominant bottlenecks (means 1.00/3 and 1.20/3; 0/50 at full maturity), primarily because (i) widely used geospatial vocabularies (CRS identifiers, ISO elements, OGC conventions) are not consistently registered in the ontology registries expected by semantic validators, and (ii) explicit, typed links to related datasets and resources are rarely exposed.

\textbf{Repository effects.}
The representative run suggests higher scores for repositories with stable identifiers and structured metadata exposure, including PANGAEA, Zenodo, Dryad, and Harvard Dataverse, and lower scores for several catalog-oriented ScienceBase and AURIN records. These are descriptive observations from an uneven, small sample rather than population estimates for each repository.

\textbf{Domain effects.}
Across domains, \textit{Interoperability} remains the principal limiting factor.
We observe a tendency for biodiversity/ecology datasets (often benefiting from established community metadata norms) to score higher overall, while datasets hosted in heterogeneous government catalog ecosystems (e.g., ScienceBase) score lower.

\subsection{Case studies: root-cause analysis}
Table~\ref{tab:case-studies} analyzes three representative datasets spanning low, medium, and high maturity.
Dimension scores are computed as the mean of sub-principles within each FAIR dimension, scaled to 0--100 (using the same rubric as Table~\ref{tab:fair-maturity-matrix}).

\textbf{AURIN OSM POIs (D2, 43.6\%).}
This dataset exhibits near-zero \textit{Interoperability} due to absent machine-readable knowledge representations and unregistered vocabularies (no JSON-LD/RDF export, no typed relations).
\textit{Accessibility} is moderate due to HTTPS access, and \textit{Reusability} is moderate due to visible license/provenance statements.
The primary gap is the lack of machine-actionable metadata exposure (structured identifiers, vocabulary grounding, and typed relations).

\textbf{NASA GDIS (D3, 71.8\%).}
This dataset achieves high \textit{Findability} via resolvable persistent identifiers and strong indexing.
\textit{Interoperability} remains moderate because metadata relies largely on generic schema markup without explicit domain vocabulary registration or typed links.
\textit{Reusability} is constrained by incomplete provenance detail and limited machine-readable linking between the dataset and related resources.

\textbf{Zenodo Crater Lake (D1, 74.4\%).}
This dataset achieves strong overall maturity, with high \textit{Reusability} driven by explicit licensing and provenance.
\textit{Interoperability} is constrained by I2/I3: the dataset relies primarily on generic vocabularies rather than registered domain vocabularies, and provides few typed links to related datasets/resources.

\subsection{Results: comparative tool analysis}

\begin{table*}[t]
\centering
\scriptsize
\setlength{\tabcolsep}{1.1pt}
\renewcommand{\arraystretch}{1.0}
\caption{Normalized dataset-level overall scores (0--100) across five FAIR assessment tools (single-run per tool). Columns D1--D50 map to the dataset list in Appendix~\ref{app:datasets}. ``--'' indicates the tool did not return a score (e.g., API timeout or FAIR-enough database limit).}
\label{tab:tool-score-matrix}
\resizebox{\textwidth}{!}{%
\begin{tabular}{@{}l*{50}{c}@{}}
\toprule
\textbf{Tool} & \textbf{D1} & \textbf{D2} & \textbf{D3} & \textbf{D4} & \textbf{D5} & \textbf{D6} & \textbf{D7} & \textbf{D8} & \textbf{D9} & \textbf{D10} & \textbf{D11} & \textbf{D12} & \textbf{D13} & \textbf{D14} & \textbf{D15} & \textbf{D16} & \textbf{D17} & \textbf{D18} & \textbf{D19} & \textbf{D20} & \textbf{D21} & \textbf{D22} & \textbf{D23} & \textbf{D24} & \textbf{D25} & \textbf{D26} & \textbf{D27} & \textbf{D28} & \textbf{D29} & \textbf{D30} & \textbf{D31} & \textbf{D32} & \textbf{D33} & \textbf{D34} & \textbf{D35} & \textbf{D36} & \textbf{D37} & \textbf{D38} & \textbf{D39} & \textbf{D40} & \textbf{D41} & \textbf{D42} & \textbf{D43} & \textbf{D44} & \textbf{D45} & \textbf{D46} & \textbf{D47} & \textbf{D48} & \textbf{D49} & \textbf{D50} \\
\midrule
\textbf{AgentFAIR} & 74.4 & 43.6 & 71.8 & 79.5 & 79.5 & 79.5 & 76.9 & 48.7 & 74.4 & 56.4 & 71.8 & 82.1 & 82.1 & 69.2 & 53.8 & 43.6 & 74.4 & 69.2 & 74.4 & 71.8 & 66.7 & 74.4 & 71.8 & 38.5 & 71.8 & 76.9 & 76.9 & 82.1 & 82.1 & 64.1 & 69.2 & 66.7 & 59.0 & 79.5 & 64.1 & 74.4 & 82.1 & 82.1 & 71.8 & 69.2 & 69.2 & 74.4 & 61.5 & 66.7 & 74.4 & 43.6 & 38.5 & 43.6 & 56.4 & 64.1 \\
F-UJI & 92.3 & 34.6 & 61.5 & 84.6 & 84.6 & 88.5 & 96.2 & 42.3 & 88.5 & 11.5 & 76.9 & 80.8 & 75.0 & 53.8 & 7.7 & 28.8 & 7.7 & 61.5 & 61.5 & 50.0 & 76.9 & 76.9 & 7.7 & 23.1 & 7.7 & 96.2 & 76.9 & 11.5 & 96.2 & 92.3 & 84.6 & 84.6 & 84.6 & 84.6 & 88.5 & 84.6 & 80.8 & 76.9 & 88.5 & 88.5 & 80.8 & 88.5 & 11.5 & 11.5 & 11.5 & 42.3 & 42.3 & 57.7 & 11.5 & 57.7 \\
FAIR-Checker & 91.7 & 54.5 & 58.3 & 91.7 & 91.7 & 91.7 & 91.7 & 20.8 & 91.7 & 75.0 & 91.7 & 91.7 & 91.7 & 91.7 & 83.3 & 50.0 & 29.2 & 62.5 & 62.5 & 45.8 & 91.7 & 91.7 & 29.2 & 58.3 & 37.5 & 79.2 & 79.2 & 79.2 & 79.2 & 79.2 & 62.5 & 62.5 & 62.5 & -- & 91.7 & 91.7 & 91.7 & 79.2 & 91.7 & 91.7 & 91.7 & 91.7 & 37.5 & 29.2 & 29.2 & 50.0 & 50.0 & 41.7 & 83.3 & 45.8 \\
FAIRshake & 100.0 & 19.4 & 69.4 & 75.0 & 80.6 & 80.6 & 91.7 & 27.8 & 88.9 & 16.7 & 94.4 & 88.9 & 86.1 & 75.0 & 16.7 & 27.8 & 36.1 & 69.4 & 72.2 & 69.4 & 88.9 & 88.9 & 36.1 & 19.4 & 36.1 & 86.1 & 80.6 & 86.1 & 91.7 & 86.1 & 63.9 & 58.3 & 58.3 & 72.2 & 75.0 & 63.9 & 69.4 & 69.4 & 88.9 & 94.4 & 91.7 & 88.9 & 36.1 & 36.1 & 36.1 & 27.8 & 27.8 & 36.1 & 16.7 & 69.4 \\
FAIR-enough & 54.5 & 40.9 & 9.1 & 54.5 & 54.5 & 50.0 & 50.0 & 45.5 & 50.0 & 22.7 & 45.5 & 50.0 & 9.1 & 18.2 & 22.7 & 18.2 & 22.7 & 68.2 & 40.9 & 9.1 & 45.5 & 45.5 & 22.7 & 36.4 & -- & 54.5 & 54.5 & 50.0 & 50.0 & 50.0 & 72.7 & 72.7 & -- & -- & -- & 45.5 & -- & -- & -- & -- & -- & -- & -- & -- & -- & -- & -- & -- & -- & -- \\
\bottomrule
\end{tabular}%
}
\end{table*}

\begin{table}[t]
\centering
\caption{Spearman association between \textsc{AgentFAIR} and baseline normalized scores.}
\label{tab:agent-baseline-correlation}
\small
\begin{tabular}{lrrr}
\toprule
\textbf{Baseline} & $\boldsymbol{n}$ & $\boldsymbol{\rho}$ & $\boldsymbol{p}$ \\
\midrule
F-UJI & 50 & 0.42 & 0.002 \\
FAIR-Checker & 49 & 0.47 & $<0.001$ \\
FAIRshake & 50 & 0.61 & $<0.001$ \\
FAIR-enough & 32 & 0.31 & 0.084 \\
\bottomrule
\end{tabular}
\end{table}

Table~\ref{tab:tool-score-matrix} compares normalized overall scores across five tools, and Table~\ref{tab:agent-baseline-correlation} reports their rank associations with \textsc{AgentFAIR}.
Across datasets, tools disagree substantially: the population standard deviation across available normalized tool scores averages 15.0 points (maximum 30.3 points), indicating that operationalizations of FAIR differ materially even after normalization.

\textbf{Failure and brittleness patterns.}
F-UJI exhibits sharp failures (scores $\leq$11.5) on 10/50 datasets (20\%), concentrated on Harvard Dataverse and Figshare, consistent with brittle identifier and registry-lookup dependencies.
FAIR-Checker timed out on one dataset (D34).
FAIR-enough produced no score for 18 datasets (36\%) due to database/runtime limits, and tends to score lowest when it does return output because its binary tests penalize partial compliance.

\textbf{Tool agreement.}
\textsc{AgentFAIR} is most aligned with FAIRshake ($\rho=0.61$, $p<0.001$, $n=50$), followed by FAIR-Checker ($\rho=0.47$, $p<0.001$, $n=49$) and F-UJI ($\rho=0.42$, $p=0.002$, $n=50$). The FAIR-enough association is weaker and not statistically significant ($\rho=0.31$, $p=0.084$, $n=32$).
This pattern reflects both rubric differences (e.g., limited \textit{Interoperability} coverage in some baseline rubrics) and extraction strategies (API-centric vs.\ browser-centric).

As a deliberately stricter diagnostic, we also compare \textsc{AgentFAIR} with the single highest baseline score available for each dataset. Under the same $\pm5$-point band, \textsc{AgentFAIR} is higher on 6/50 datasets, comparable on 8/50, and lower on 36/50. This ``best baseline'' is a composite oracle rather than one deployable tool; the result is reported to prevent the average-baseline analysis from being misread as a superiority claim.

Qualitatively, three system properties contribute to differences:

\begin{itemize}
    \item \textbf{Contextual resolution of identifiers and access conditions:} LLM agents can disambiguate DOI-like strings, reconcile multiple identifiers, and interpret whether authentication statements are optional or mandatory.
    \item \textbf{Geospatial sensitivity:} Prompts explicitly recognize CRS identifiers, ISO metadata elements, and OGC endpoint mentions, which informs \textit{Interoperability} and R1.3 scoring.
    \item \textbf{Critic-based reconciliation:} Cross-sub-principle checks trigger re-evaluation when evidence suggests inconsistency (e.g., protocol openness vs.\ authentication clarity).
\end{itemize}

\subsection{Results: cost and efficiency analysis}

\begin{figure}[t]
    \centering
    \includegraphics[width=0.6\columnwidth]{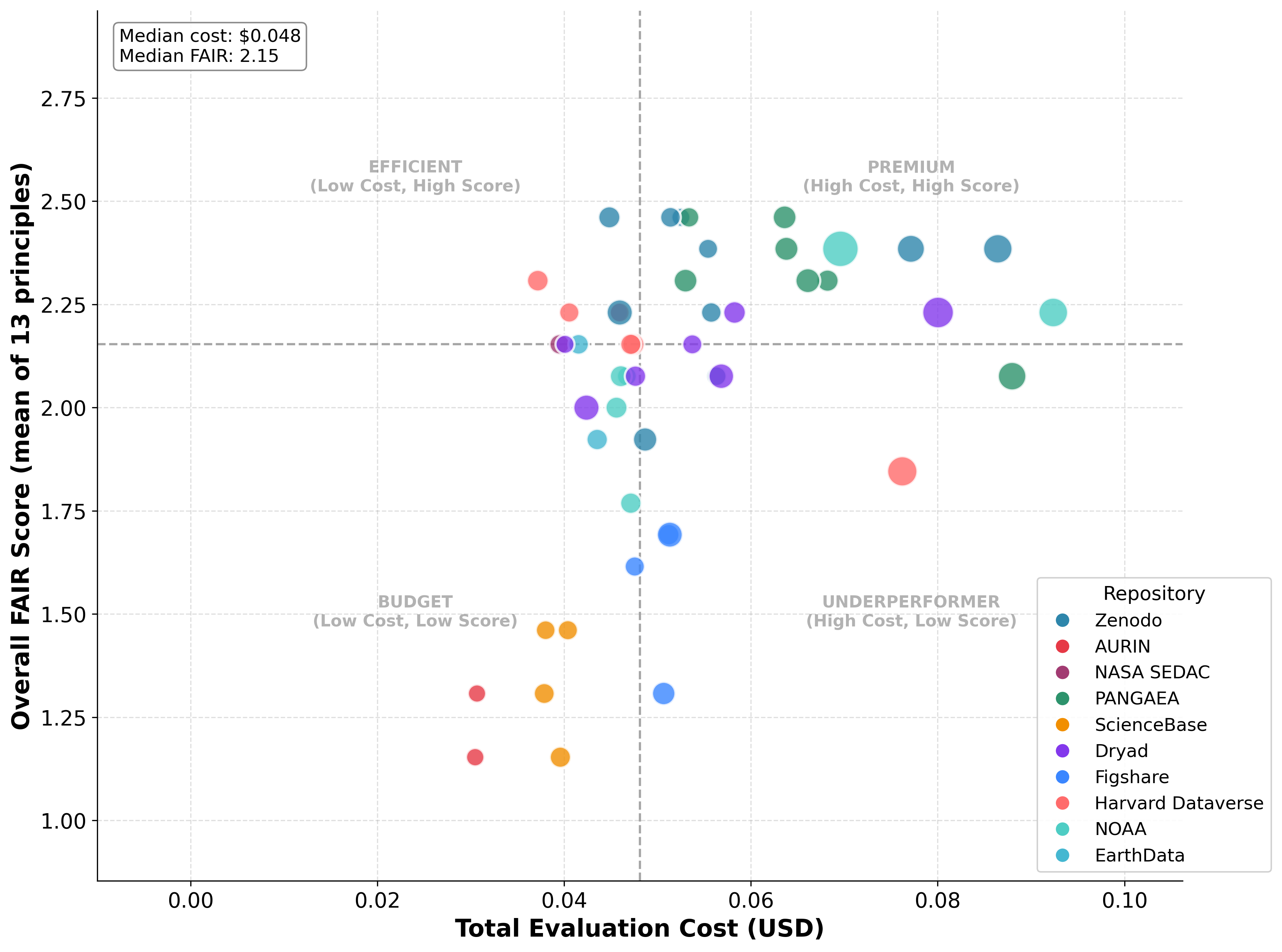}
    \Description{Scatter plot positioning datasets by evaluation cost versus overall FAIR score, highlighting low-cost high-score clusters.}
    \caption{Cost-efficiency quadrant: datasets positioned by evaluation cost (x-axis) vs.\ overall FAIR score (y-axis).}
    \label{fig:cost-quadrant}
\end{figure}

\begin{table}[t]
\centering
\caption{\textsc{AgentFAIR} efficiency metrics across 50 datasets}
\label{tab:efficiency}
\begin{tabular}{lr}
\toprule
\textbf{Metric} & \textbf{Value} \\
\midrule
Total Tokens & 16.54M \\
Total Cost & \$2.68 \\
Mean Cost/Dataset & \$0.054 \\
Mean Processing Time & 1{,}054 seconds \\
\bottomrule
\end{tabular}
\end{table}

Figure~\ref{fig:cost-quadrant} and Table~\ref{tab:efficiency} summarize computational characteristics.
Zenodo datasets cluster in a ``low cost / high score'' region, reflecting stable structured metadata exposure.
More complex repositories and datasets require additional crawling and semantic reasoning, increasing processing time and token usage.
A cost breakdown by FAIR dimension appears in Appendix (Figure~\ref{fig:cost-breakdown}).

\subsection{Consistency and auditability analysis}
\label{sec:consistency}
Across repeated evaluations during development (5 runs per dataset on a 10-dataset subset, stratified by FAIR maturity quintile), mean exact agreement on sub-principle scores is 89\%, with a reported run-level standard deviation of 3 percentage points. We omit the earlier confidence interval because the retained summary statistics are insufficient to reconstruct it correctly. This analysis measures repeatability, not reproducibility in the deterministic sense. AgentFAIR also retains deterministic extraction failure modes; browser rendering and retry logic reduce some failures but do not eliminate them.

\textbf{Failure analysis.}
Of 50 datasets, 3 (6\%) required manual URL correction due to redirect chains or domain changes.
The critic triggered re-evaluation in 96\% of sub-principle assessments. A trigger means that at least one hard logical check failed or confidence fell below its threshold; it does not by itself show that the initial score was wrong or that the revised score is more accurate.
In contrast, F-UJI failed completely on 10 datasets (20\%) due to identifier resolution errors, and FAIR-enough encountered database errors on 18 datasets (36\%).

\textbf{Human evaluation protocol.}
Three domain experts with 5+ years of geospatial data management experience independently evaluated a stratified random sample of 15 datasets (3 per maturity quintile).
After a calibration session with worked rubric examples, each expert independently assessed all 13 sub-principles without access to automated tool outputs (195 judgments per expert; 585 annotations in total). Per-dataset annotation time ranged from a few minutes for simple records to roughly 20 minutes for complex cases.
Inter-rater agreement (Fleiss' $\kappa$=0.71, 95\% CI: [0.65, 0.77]) indicates substantial agreement.
\textsc{AgentFAIR} achieved 82\% alignment with expert consensus (majority vote); disagreements concentrate on I2 (interpreting vocabulary FAIRness, $\kappa$=0.58) and R1.2 (provenance completeness thresholds, $\kappa$=0.61), indicating where rubric definitions can be further refined.

\textbf{Baseline comparison on the expert subset.}
The baseline tools do not expose outputs directly compatible with all 13 ordinal expert labels, so a head-to-head sub-principle accuracy comparison is not available. As a coarser diagnostic, we aggregate the expert labels to an overall score and apply a $\pm5$-point dataset-level band. F-UJI and FAIR-Checker each fall within the band on 2/15 datasets, FAIRshake on 4/15, and FAIR-enough on 0/10 returned outputs (five failures). These figures are not commensurate with \textsc{AgentFAIR}'s 82\% sub-principle alignment and should not be interpreted as proof of superior accuracy. A larger study with tool outputs mapped and judged at the same sub-principle level remains necessary.

\subsection{Within-family model transfer}
To probe dependence on GPT-4o-mini, we reran the identical pipeline and prompts with GPT-4o on the same stratified 10-dataset subset. Dataset rankings are strongly associated ($\rho=0.877$, $p<0.001$); 92/130 sub-principle scores are exact matches and 121/130 (93.1\%) differ by at most one maturity level. The mean absolute difference in normalized score is 7.1 percentage points. This result is supportive within the GPT family, but it does not establish robustness to model version changes or transfer to a different model family.

\subsection{Ablation study: critic agent contribution}

\begin{table}[t]
\centering
\caption{Ablation: critic agent contribution ($n$=10 datasets)}
\label{tab:ablation}
\small
\begin{tabular}{lcc}
\toprule
\textbf{Configuration} & \textbf{Consistency} & \textbf{Re-eval Rate} \\
\midrule
Full system (with critic) & 89$\pm$3\% & 96\% triggered \\
Without critic agent & 71$\pm$5\% & N/A \\
\bottomrule
\end{tabular}
\par\smallskip
{\footnotesize Consistency = agreement across 5 repeated runs; Re-eval Rate = percentage of sub-principle assessments triggering critic review.}
\end{table}

To isolate the critic agent's contribution, we conducted an ablation study on 10 representative datasets (2 from each FAIR maturity quintile, spanning Zenodo, Dryad, PANGAEA, Harvard Dataverse, and NOAA NCEI).
Table~\ref{tab:ablation} compares full \textsc{AgentFAIR} against a variant with the critic agent disabled.
Without critic-based quality control, consistency drops from 89$\pm$3\% to 71$\pm$5\% ($p<0.01$, paired $t$-test), with the largest degradation on I2 and R1.2---sub-principles requiring nuanced evidence interpretation.
The 96\% trigger rate combines seven hard logical checks with confidence gates. The experiment did not retain a trigger-type breakdown, conditional score-change distribution, latency contribution, or accuracy of the intermediate outputs; consequently, the ablation supports a repeatability benefit but does not fully characterize whether the critic corrects genuine errors. 

\subsection{Limitations and future experiments}
The study is exploratory and has several unresolved validity limits. First, 50 datasets and 15 expert-annotated datasets do not support broad claims across the long tail of geospatial repositories. Second, the cross-tool normalization conflates different constructs and is retained only for disagreement analysis; baseline runtimes were not systematically logged. Third, the current evidence does not isolate browser engineering, LLM enrichment, geospatial prompt terms, multi-agent decomposition, and critic retries. Geospatial-prompt, rule-only extraction, temperature 0.0 versus 0.1, and plain-retry ablations remain outstanding. Fourth, GPT-4o is a within-family check rather than cross-family validation, and proprietary model updates threaten longitudinal repeatability. Fifth, the strict I2 registry criterion can under-credit valid community vocabularies absent from generic registries; the low expert agreement on I2 reinforces this concern. Sixth, prompt-injection and poisoned-metadata robustness have not been evaluated. Finally, raw repeated-run, human-label, ablation, and model-transfer outputs should be released with analysis scripts before the reported inferential statistics are treated as independently reproducible.

\section{Conclusion}
We introduced \textsc{AgentFAIR}, an LLM-driven multi-agent framework for auditable FAIRness evaluation of geospatial datasets. The system combines browser-based and structured metadata extraction, 13 rubric-specific evaluators, deterministic checks, and critic-guided retries.

In a 50-dataset, 10-repository sample, \textit{Findability} is strongest (79.7\%) and \textit{Interoperability} weakest (45.3\%). Cross-tool normalized scores show substantial disagreement (mean per-dataset standard deviation 15.0 points), but heterogeneous rubrics prevent interpreting those differences as comparative accuracy. Preliminary repeated-run and expert studies support further investigation of the critic and evidence model, while the high retry rate, small human sample, missing component ablations, and within-family-only model check leave important questions unresolved. At approximately \$0.054 per dataset, the framework is economically feasible for larger validation studies; stronger accuracy or deployment claims require released raw evaluation artifacts, cross-family testing, and adversarial security evaluation.

\bibliographystyle{ACM-Reference-Format}
\bibliography{references}


\appendix

\section{Open-source artifact and prompt specifications}
\label{app:artifact}

\subsection{Zenodo release and contents}
An archived software snapshot is available on Zenodo
(\url{https://doi.org/10.5281/zenodo.18529560}); the maintained source is intended for
\url{https://github.com/MingCHEN-Github/AgentFAIR}.
The Zenodo deposit contains:
\texttt{AgentFAIR.zip} (full source code),
\texttt{DESCRIPTION.md} (setup + artifact overview),
and \texttt{app\_demo.mp4} (end-to-end walkthrough video).
The software is licensed under AGPL-3.0-only.

\subsection{Quick start and environment configuration}
\noindent\textbf{Prerequisites.}
Python 3.10+ and an OpenAI API key. The archived artifact is self-contained (no git required).

\noindent\textbf{One-command launch.}
\begin{quote}\small\ttfamily
unzip AgentFAIR.zip\\
cd AgentFAIR\\
./run.sh
\end{quote}

\noindent\texttt{run.sh} automates setup and execution, including dependency installation,
Playwright browser installation for web crawling, creation of \texttt{.env} from
\texttt{.env.example} when needed, launching the FastAPI backend
(\texttt{http://localhost:8000}), and opening the React/Vite frontend in a browser.

\noindent\textbf{API key configuration.}
Keys can be provided either (i) via the frontend \emph{Settings} modal at runtime
(recommended for quick evaluation) or (ii) via a local \texttt{.env} file.
The artifact also supports optional keys for Brave Search (\texttt{BRAVE\_API\_KEY}),
OpenRouter (\texttt{OPENROUTER\_API\_KEY}), and opt-in LangSmith tracing
(\texttt{LANGSMITH\_API\_KEY}). Remote tracing is disabled by default.

\subsection{Repository layout and persisted outputs}
Unzipping \texttt{AgentFAIR.zip} produces an \texttt{AgentFAIR/} directory. Key components:
\begin{quote}\small\ttfamily
fair\_agents/ \ \ \ \ \ core Python package (agents, tools, LangGraph workflows, reporting)\\
fair\_agents/api/ \ \ FastAPI service (local evaluation endpoints)\\
frontend/ \ \ \ \ \ \ React/Vite client (run orchestration + results UI)\\
storage/ \ \ \ \ \ \ \ generated SQLite evidence store (ignored by Git)\\
fair\_agents/reports/ \ generated Markdown/JSON reports (ignored by Git)\\
tests/ \ \ \ \ \ \ \ URL-safety regression tests\\
run.sh \ \ \ \ \ \ \ \ \ single-command launcher
\end{quote}

\noindent\textbf{Audit trail outputs.}
Each evaluation persists (i) per-sub-principle scores and confidence,
(ii) evidence bundles with provenance pointers/snippets,
and (iii) exported reports (Markdown + JSON) backed by a SQLite evidence store.
Optional LangSmith traces can be enabled when a tracing key is provided.

\subsection{Representative prompt template and output schema}
\label{app:prompts}

\textsc{AgentFAIR} prompts follow a uniform multi-part structure:
(i) sub-principle definition and scope,
(ii) maturity rubric (0--3; operationalization in Table~2),
(iii) decision logic and edge cases,
(iv) geospatial adaptations (when applicable),
(v) evidence requirements,
(vi) anti-hallucination constraints,
(vii) output JSON schema, and
(viii) short worked examples.

Below we show a simplified \emph{illustrative} template for a sub-principle agent.
In practice, each agent instantiates this template with the corresponding rubric criteria
and geospatial indicators from the operationalization in Table~2.

\begin{quote}\small\ttfamily
\textbf{Input:} extracted metadata record M (structured fields + landing page snippets).\\
\textbf{Task:} Evaluate sub-principle \{F1/F2/...\} using the rubric below.\\
\textbf{Rubric:} score 0/1/2/3 with explicit criteria.\\
\textbf{Evidence rule:} every claim must cite a field or quoted snippet from M.\\
\textbf{Output JSON:}\\
\{\\
\ \ "principle": "...",\\
\ \ "score": 0|1|2|3,\\
\ \ "confidence": 0.0--1.0,\\
\ \ "evidence": [\{"source": "...", "snippet": "..."\}, ...],\\
\ \ "rationale": "...",\\
\ \ "recommendations": ["...", ...]\\
\}
\end{quote}

\section{Further Discussions of \textsc{AgentFAIR}}

\subsection{Key insights}
\textbf{\textit{Interoperability} is the dominant bottleneck.}
Across repositories, I2 and I3 consistently score lowest.
A key driver is that widely used geospatial vocabularies and standards are not consistently represented in the registries relied upon by semantic FAIR validators, which leads to systematic penalties.
This suggests that improving ecosystem-level vocabulary registration (not only dataset-level metadata) is critical for improving geospatial FAIRness.

\textbf{Rule-based tools are brittle on modern repository interfaces.}
We observe sharp baseline failures when crawlers cannot access machine-readable metadata (e.g., JavaScript-rendered pages) or when identifier resolution relies on assumptions that do not hold universally.
These failures can dominate overall tool disagreement and complicate the interpretation of FAIR scores as decision signals.

\textbf{LLM-based evaluation can be useful but must be auditable.}
LLM agents help interpret ambiguous natural language and recover domain cues, but they introduce stochasticity.
In \textsc{AgentFAIR}, evidence-grounded prompts, explicit rubrics, and critic-based consistency checks are essential to make LLM reasoning usable for compliance assessment.

\subsection{Threats to validity}
\textbf{Dataset sampling.}
Our benchmark emphasizes cross-repository diversity, but 50 datasets is not exhaustive; results may shift under different sampling strategies or when focusing on a single community.

\textbf{Baseline normalization.}
Normalizing heterogeneous tool outputs to a single 0--100 score enables comparison but necessarily abstracts away rubric differences.
We therefore interpret correlations and variance as indicators of \emph{disagreement}, not as absolute correctness.

\textbf{Lack of ground-truth FAIR labels.}
FAIRness is not a single objective quantity; expert judgments may differ, and community standards evolve.
We address this only partially by reporting evidence and documenting disagreements; the preliminary expert sample is not a definitive ground truth.

\textbf{LLM-specific threats.}
The primary evaluation relies on GPT-4o-mini, which may change without a versioned model snapshot. The GPT-4o check provides only within-family evidence. Low temperature does not guarantee determinism, and no temperature ablation was run. Prompt phrasing can affect score distributions, and the prompts were iteratively refined during development. Evidence requirements and critic checks improve auditability but cannot prevent hallucination, prompt injection, or subtle misinterpretation of poisoned or ambiguous metadata. Repository-specific overfitting therefore remains possible.

\textbf{I2 construct validity.}
The experimental I2 rubric places substantial weight on vocabulary registry presence. Valid geospatial vocabularies may be documented and accepted in community practice without appearing in generic registries, so the rubric can systematically under-credit them. Future work should combine registry evidence with community governance, resolvability, versioning, licensing, and documentation rather than treating registry count as a normative definition of FAIRness.

\subsection{Applications}
\textsc{AgentFAIR} supports: (i) \textbf{repository FAIRification} (prioritized remediation roadmaps), (ii) \textbf{CI/CD integration} (continuous metadata quality monitoring), (iii) \textbf{funding compliance reporting} (evidence-backed audit trails), and (iv) \textbf{training data curation} (filtering and ranking datasets for geospatial foundation models based on FAIR maturity).



\section{LangGraph workflow during evaluation process}
\begin{figure*}[h]
    \centering
    \includegraphics[width=0.85\textwidth]{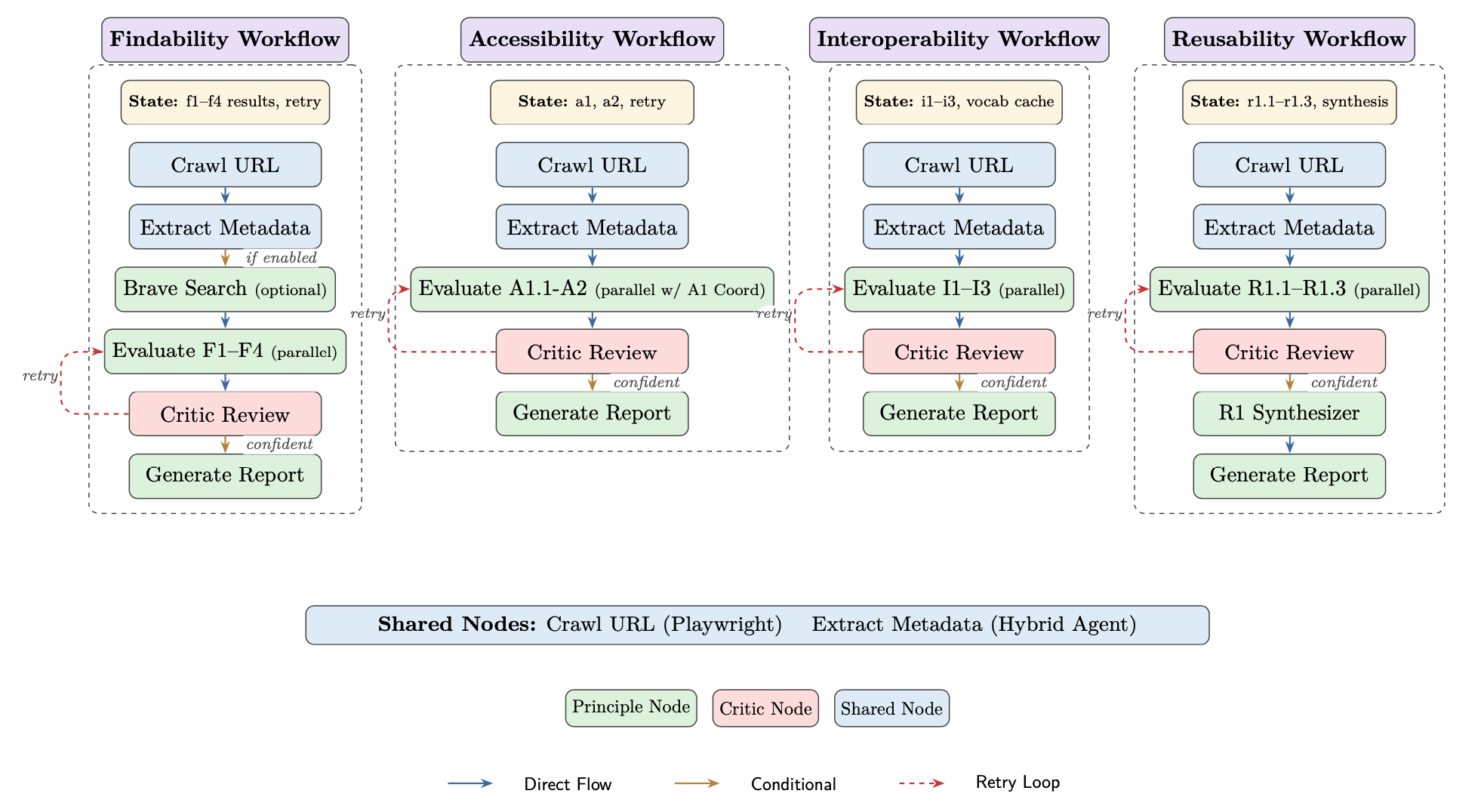}
    \Description{Workflow diagram showing four FAIR-dimension subgraphs with crawl, extract, evaluate, critic, and report nodes, including retry loops.}
    \caption{LangGraph state machine orchestrating FAIR evaluation workflows with conditional edges and retry logic.}
    \label{fig:LangGraph}
\end{figure*}

The proposed \textsc{AgentFAIR} employs LangGraph, an open-source orchestration library in the LangChain ecosystem~\cite{LangGraph}, to design multi-agent workflows as directed state machines. As illustrated in Figure ~\ref{fig:LangGraph}, each FAIR dimension operates as an independent graph composed of modular nodes—crawling, metadata
extraction, principle-specific evaluation, critic review, and reporting—connected
through conditional and retry-based edges. Shared state variables capture crawl
results, agent confidence, and error counts, enabling traceable and auditable
execution.
\label{app:langgraph}

\section{Additional plots and tables}

\begin{figure}[htbp]
    \centering
    \includegraphics[width=\columnwidth]{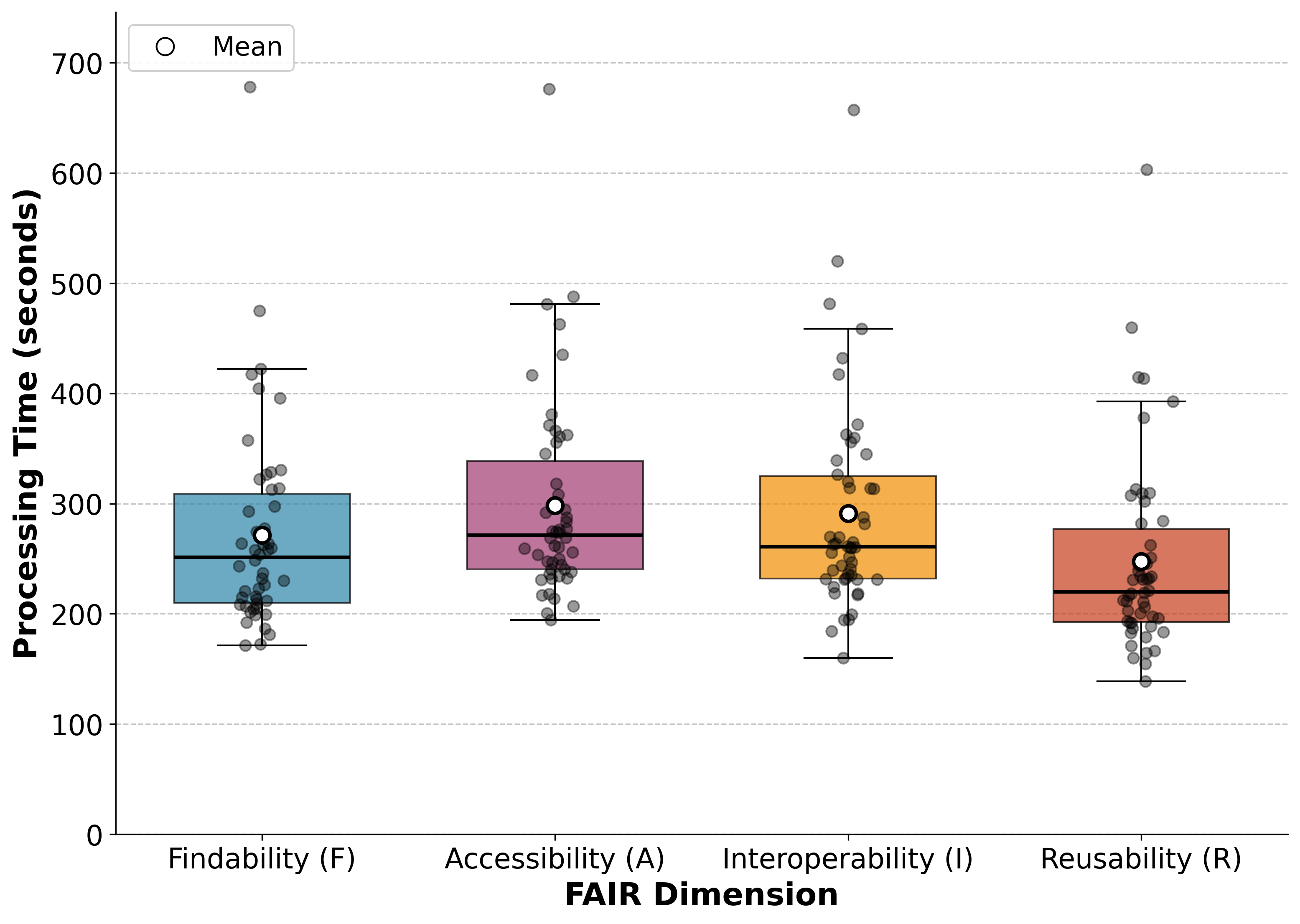}
    \Description{Boxplots of processing time by FAIR dimension across datasets.}
    \caption{Processing time distribution by FAIR dimension.}
    \label{fig:time-boxplot}
\end{figure}

Figure~\ref{fig:time-boxplot} shows that processing time varies by FAIR dimension; \textit{Reusability} is fastest, while \textit{Findability} and \textit{Accessibility} take longest due to crawling and metadata extraction overhead.
Most evaluations cluster in the 150--350 second range per dimension, with outliers corresponding to metadata-heavy datasets.

\begin{figure}[htbp]
    \centering
    \includegraphics[width=\columnwidth]{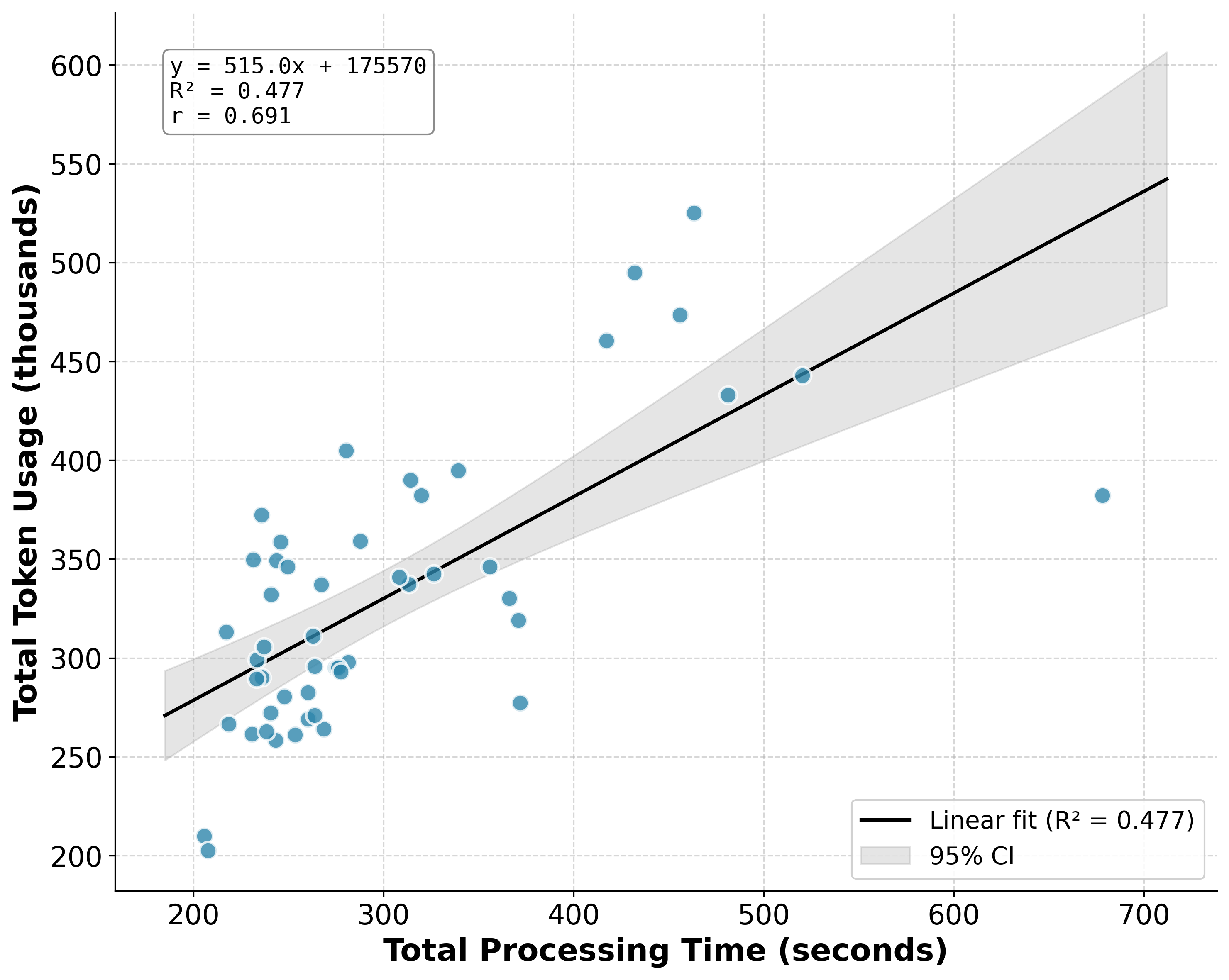}
    \Description{Scatter plot showing the relationship between token usage and processing time.}
    \caption{Token usage vs.\ processing time correlation.}
    \label{fig:token-scatter}
\end{figure}

Figure~\ref{fig:token-scatter} reveals a strong positive correlation between token usage and processing time, confirming that LLM API calls dominate wall-clock time.
This relationship suggests that token-reduction strategies can directly lower both evaluation cost and latency.

\begin{figure*}[htbp]
    \centering
    \includegraphics[width=\textwidth]{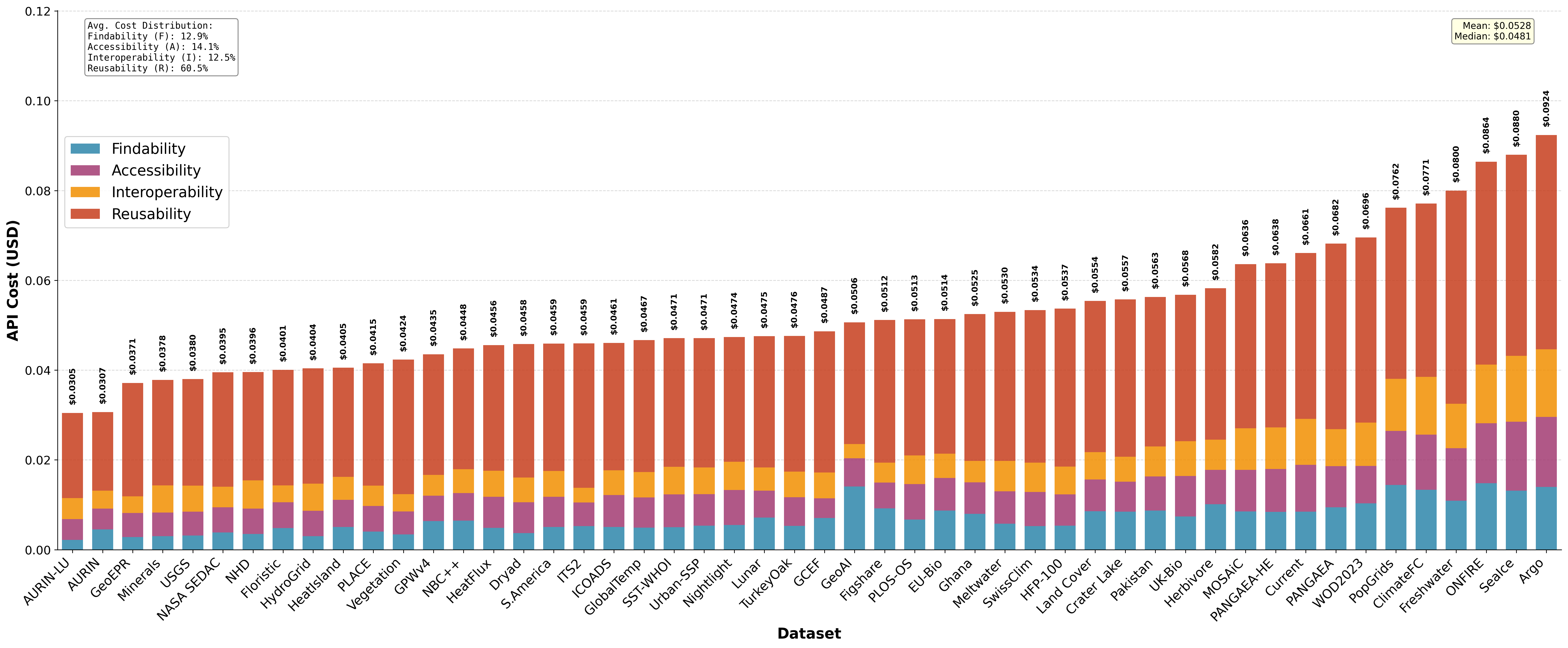}
    \Description{Bar chart breaking down API cost by FAIR dimension.}
    \caption{API cost breakdown by FAIR dimension across 50 datasets.}
    \label{fig:cost-breakdown}
\end{figure*}

Figure~\ref{fig:cost-breakdown} sorts datasets by total cost (ascending), with \textit{Reusability} consistently contributing the largest cost share and \textit{Interoperability} the smallest.
Total evaluation cost for all 50 datasets remains under \$3, demonstrating the economic viability of LLM-based FAIR assessment at scale.

\begin{figure}[htbp]
    \centering
    \includegraphics[width=\columnwidth]{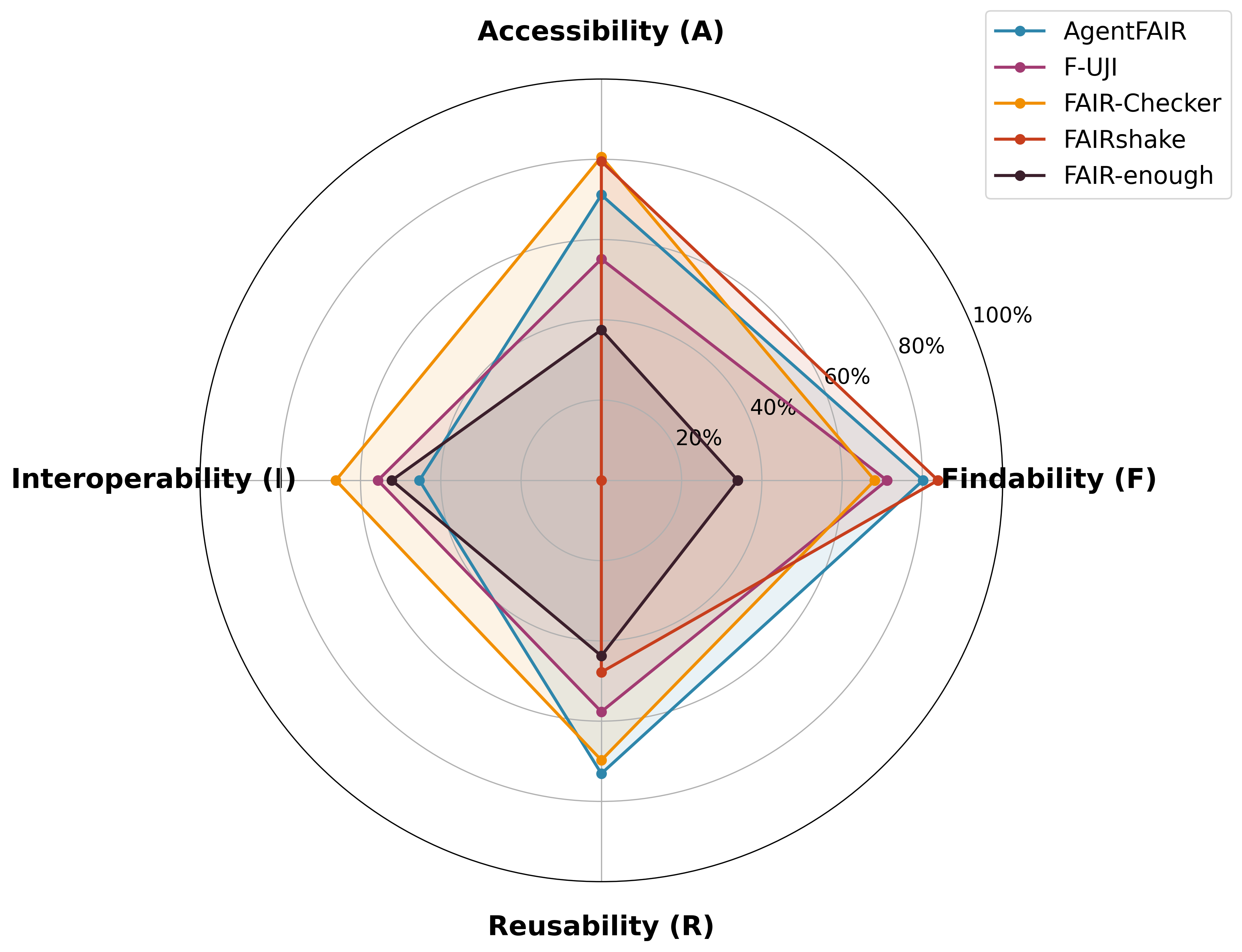}
    \Description{Radar plot comparing tools across FAIR dimensions and methodological attributes.}
    \caption{FAIR dimension coverage radar chart across tools.}
    \label{fig:radar}
\end{figure}

Figure~\ref{fig:radar} compares mean dimension-level scores across five FAIR assessment tools.
AgentFAIR and F-UJI show relatively balanced profiles across all four dimensions, whereas FAIRshake lacks \textit{Interoperability} coverage entirely and FAIR-enough tends to score lower due to its binary test methodology.

\begin{figure}[h]
    \centering
    \includegraphics[width=\columnwidth]{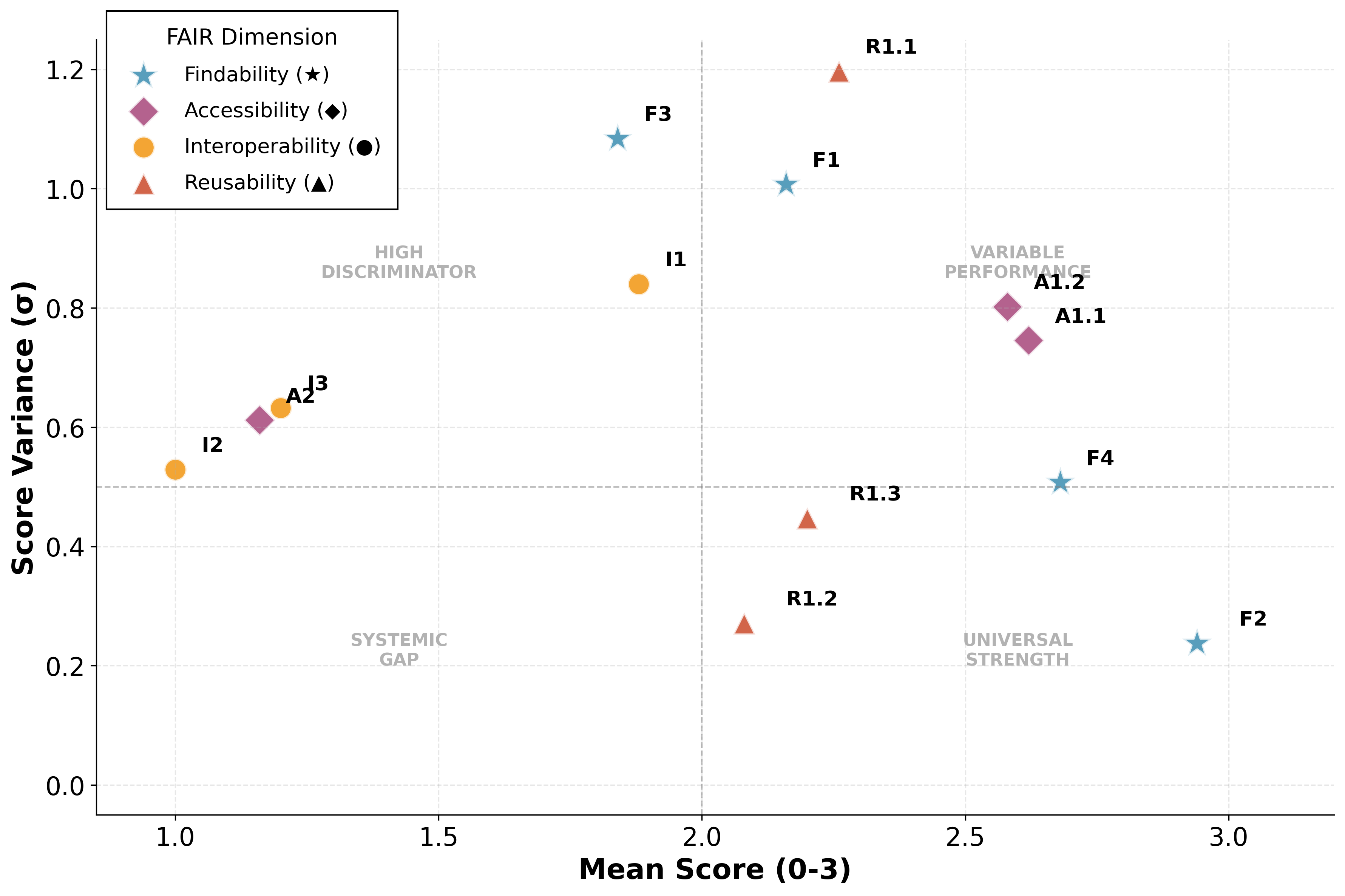}
    \Description{Scatter plot summarizing each FAIR principle by mean maturity and variance across datasets.}
    \caption{FAIR principle maturity landscape: mean score vs. variance across datasets.}
    \label{fig:maturity}
\end{figure}

\begin{figure*}[htbp]
    \centering
    \includegraphics[width=\textwidth]{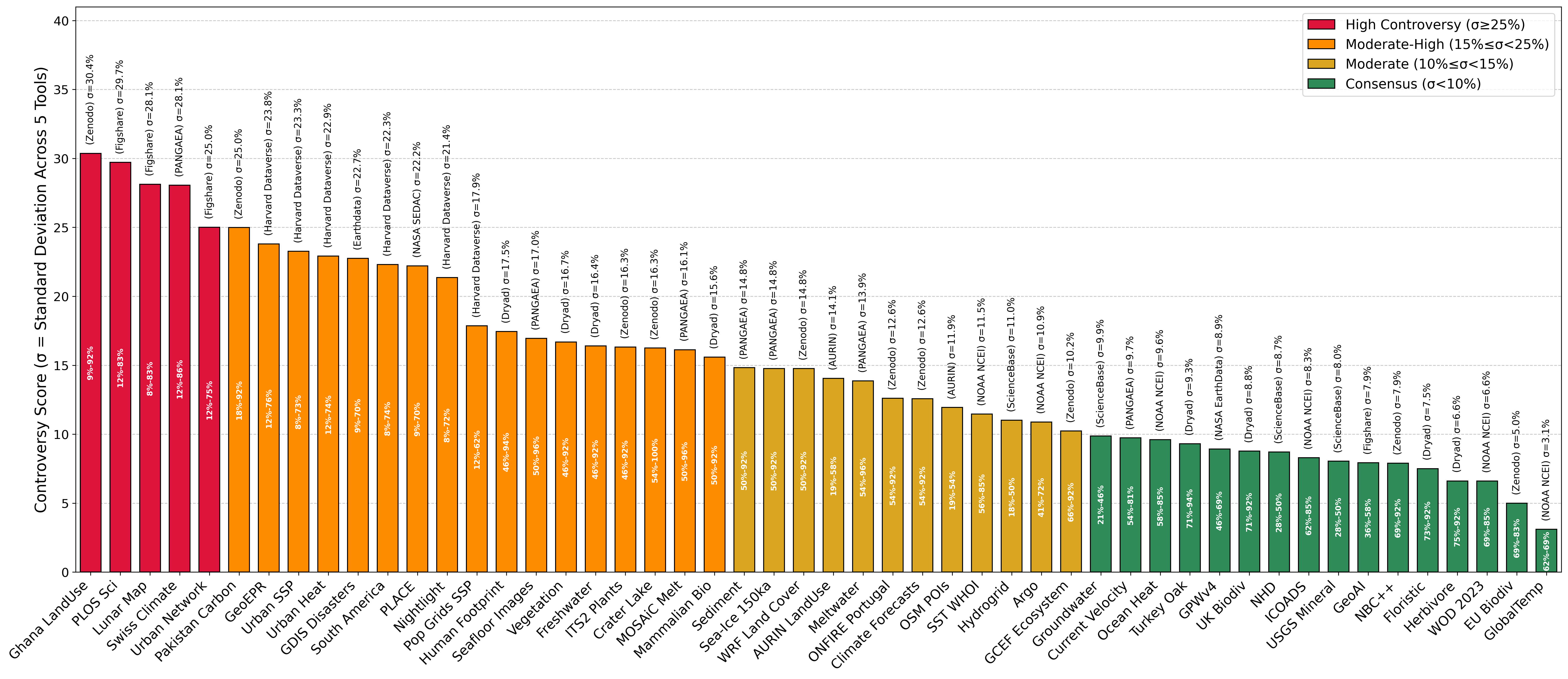}
    \Description{Bar chart showing per-dataset score variance across tools, labeled as a controversy index.}
    \caption{Dataset controversy index: scoring variance across tools.}
    \label{fig:controversy}
\end{figure*}

Figure~\ref{fig:controversy} ranks datasets by scoring variance (standard deviation) across tools, with high-controversy datasets exhibiting >25\% standard deviation.
These datasets expose fundamental differences in how each tool operationalizes FAIR principles and serve as valuable diagnostic benchmarks for tool comparison.

Figure~\ref{fig:maturity} plots each FAIR sub-principle by its mean maturity score and cross-dataset variance.
\textit{Findability} and \textit{Accessibility} principles cluster in the ``Universal Strength'' quadrant (high mean, low variance), while \textit{Interoperability} sub-principles fall into the ``High Discriminator'' or ``Systemic Gap'' regions, confirming that I-dimension compliance is the most variable and challenging aspect of FAIR maturity.

\section{Dataset details}
Table~\ref{tab:dataset-full} summarizes the key detailed information for all the evaluated 50 datasets.
\label{app:datasets}
\begin{table*}[h]
\centering
\caption{Full dataset summary (50 datasets, 10 repositories).}
\label{tab:dataset-full}

\scriptsize
\setlength{\tabcolsep}{3.5pt}
\renewcommand{\arraystretch}{1.05}

\begin{tabularx}{\textwidth}{r Y l Y r Y l Y}
\toprule
\textbf{\#} & \textbf{Dataset} & \textbf{Repository} & \textbf{Domain} &
\textbf{\#} & \textbf{Dataset} & \textbf{Repository} & \textbf{Domain} \\
\midrule
D1  & Crater Lake       & Zenodo            & Marine/\allowbreak Geoscience &
D26 & Meltwater         & PANGAEA           & Climate/\allowbreak Hydrology \\
D2  & OSM POIs          & AURIN             & Urban/\allowbreak Land Classification &
D27 & Current Velocity  & PANGAEA           & Marine/\allowbreak Geoscience \\
D3  & GDIS Disasters    & Earthdata         & Disaster/\allowbreak Emissions &
D28 & Swiss Climate     & PANGAEA           & Climate/\allowbreak Hydrology \\
D4  & ONFIRE Portugal   & Zenodo            & Disaster/\allowbreak Emissions &
D29 & MOSAiC Melt       & PANGAEA           & Marine/\allowbreak Geoscience \\
D5  & Climate Forecasts & Zenodo            & Climate/\allowbreak Hydrology &
D30 & Sea-Ice 150ka     & PANGAEA           & Marine/\allowbreak Geoscience \\
D6  & WRF Land Cover    & Zenodo            & Urban/\allowbreak Land Classification &
D31 & ICOADS            & NOAA NCEI         & Marine/\allowbreak Geoscience \\
D7  & Seafloor Images   & PANGAEA           & Marine/\allowbreak Geoscience &
D32 & Ocean Heat        & NOAA NCEI         & Climate/\allowbreak Hydrology \\
D8  & Groundwater       & ScienceBase       & Climate/\allowbreak Hydrology &
D33 & SST WHOI          & NOAA NCEI         & Marine/\allowbreak Geoscience \\
D9  & Mammalian Bio     & Dryad             & Biodiversity/\allowbreak Ecology &
D34 & WOD 2023          & NOAA NCEI         & Marine/\allowbreak Geoscience \\
D10 & Urban Network     & Figshare          & Urban/\allowbreak Land Classification &
D35 & GCEF Ecosystem    & Zenodo            & Biodiversity/\allowbreak Ecology \\
D11 & Human Footprint   & Dryad             & Biodiversity/\allowbreak Ecology &
D36 & ITS2 Plants       & Zenodo            & Biodiversity/\allowbreak Ecology \\
D12 & Sediment          & PANGAEA           & Marine/\allowbreak Geoscience &
D37 & NBC++             & Zenodo            & Biodiversity/\allowbreak Ecology \\
D13 & Ghana LandUse     & Zenodo            & Urban/\allowbreak Land Classification &
D38 & EU Biodiv         & Zenodo            & Biodiversity/\allowbreak Ecology \\
D14 & Pakistan Carbon   & Zenodo            & Disaster/\allowbreak Emissions &
D39 & Floristic         & Dryad             & Biodiversity/\allowbreak Ecology \\
D15 & Lunar Map         & Figshare          & Planetary Science &
D40 & Turkey Oak        & Dryad             & Biodiversity/\allowbreak Ecology \\
D16 & Hydrogrid         & ScienceBase       & Climate/\allowbreak Hydrology &
D41 & UK Biodiv         & Dryad             & Biodiversity/\allowbreak Ecology \\
D17 & South America     & Harvard Dataverse & Regional Analysis &
D42 & Herbivore         & Dryad             & Biodiversity/\allowbreak Ecology \\
D18 & GlobalTemp        & NOAA NCEI         & Climate/\allowbreak Hydrology &
D43 & Pop Grids SSP     & Harvard Dataverse & Demographics \\
D19 & Argo              & NOAA NCEI         & Marine/\allowbreak Geoscience &
D44 & Urban Heat        & Harvard Dataverse & Urban/\allowbreak Land Classification \\
D20 & PLACE             & NASA SEDAC        & Demographics &
D45 & GeoEPR            & Harvard Dataverse & Regional Analysis \\
D21 & Vegetation        & Dryad             & Biodiversity/\allowbreak Ecology &
D46 & USGS Mineral      & ScienceBase       & Geology \\
D22 & Freshwater        & Dryad             & Biodiversity/\allowbreak Ecology &
D47 & NHD               & ScienceBase       & Climate/\allowbreak Hydrology \\
D23 & Nightlight        & Harvard Dataverse & Remote Sensing &
D48 & GeoAI             & Figshare          & Technology \\
D24 & AURIN LandUse     & AURIN             & Urban/\allowbreak Land Classification &
D49 & PLOS Sci          & Figshare          & Scientific Publishing \\
D25 & Urban SSP         & Harvard Dataverse & Demographics &
D50 & GPWv4             & NASA EarthData    & Demographics \\
\bottomrule
\end{tabularx}
\end{table*}

\section{Artifact and auditability checklist}
\begin{itemize}
    \item \textbf{Hardware:} Apple M2 Pro, 32\,GB RAM, macOS 15
    \item \textbf{Software:} Python 3.11; current release requires Python 3.10+, LangChain Core 1.x, and LangGraph 1.x
    \item \textbf{Model:} GPT-4o-mini via OpenAI API (temperature=0.1)
    \item \textbf{Dependencies:} Playwright, FastAPI, SQLite, Extruct, lxml, and rdflib
    \item \textbf{Logging:} prompts, responses, evidence, and scores are persisted as an audit trail
    \item \textbf{Code:} \url{https://github.com/MingCHEN-Github/AgentFAIR}; archived snapshot at \url{https://doi.org/10.5281/zenodo.18529560}
    \item \textbf{Missing research artifacts:} raw repeated-run outputs, expert labels, ablation runs, model-transfer outputs, and analysis notebooks are not included in the current software checkout and are required for independent reproduction of the reported statistics
\end{itemize}

\clearpage
\onecolumn
\section{Comparative analysis of FAIR assessment systems}

\begin{center}
    \captionof{table}{Comparative analysis of FAIR assessment systems and this work's contributions}
    \label{tab:lit-comparison}
    \small
    \renewcommand{\arraystretch}{1.2}
    \setlength{\tabcolsep}{4pt}
    \rowcolors{2}{gray!10}{white}

    \resizebox{\textwidth}{!}{%
    \begin{tabular}{|l|c|c|c|c|c|c|}
    \hline
    \rowcolor{gray!25}
    \textbf{System / Study} &
    \textbf{LLM-Based} &
    \textbf{Geospatial} &
    \textbf{Explainable} &
    \textbf{Traceable} &
    \textbf{Cross-Principle} &
    \textbf{All 13 Sub-Principles} \\
    \hline

    F-UJI \citep{devaraju2020fujifair} & \xmark & \xmark & \cmark & \cmark & \xmark & \cmark \\
    FAIRshake \citep{clarke2019fairshake} & \xmark & \xmark & \cmark & \cmark & \xmark & \xmark \\
    AutoDCWorkflow \citep{li2024autodcworkflow} & \cmark & \xmark & \cmark & \xmark & \xmark & \xmark \\
    TKFDM Tool \citep{tkfdm2023fairtool} & \xmark & \xmark & \cmark & \cmark & \xmark & \xmark \\
    FAIR-Checker \citep{gaignard2023fair} & \xmark & \xmark & \cmark & \cmark & \cmark & \xmark \\
    FAIR Evaluator \citep{fair_evaluator_cookbook} & \xmark & \xmark & \cmark & \cmark & \xmark & \cmark \\
    FAIR-enough \citep{fair_enough} & \xmark & \xmark & \cmark & \cmark & \xmark & \cmark \\
    Cocoon \citep{huang2024cocoon} & \cmark & \xmark & \cmark & \cmark & \xmark & \xmark \\
    FAIRBridge \citep{sakib2025fairbridge} & \cmark & \xmark & \cmark & \cmark & \xmark & \xmark \\
    AgReFed \citep{bahlo2024agrefed_dsj} & \xmark & \cmark & \cmark & \cmark & \xmark & \xmark \\
    OGC Disaster Pilot \citep{ogc_iso_sprint2022} & \cmark & \cmark & \cmark & \cmark & \xmark & \xmark \\
    \textbf{This Work} & \cmark & \cmark & \cmark & \cmark & \cmark & \cmark \\
    \hline
    \end{tabular}%
    }
\end{center}

\section{Diagnostic comparison with the baseline average}
\label{app:detailed-comparison}

For completeness, averaging the available baseline scores per dataset and applying a $\pm5$-point band places \textsc{AgentFAIR} above that average on 20/50 datasets, within the band on 18/50, and below it on 12/50. This statistic is sensitive to missing baseline outputs and mixes non-equivalent rubrics. It is therefore retained only as a descriptive disagreement summary; the stricter best-baseline result is reported in Section~5, and neither comparison is treated as an accuracy ranking. We omit the earlier per-dataset ``higher/lower'' narratives because many large differences were attributable to crawler and redirect failures rather than the LLM or multi-agent design.

\end{document}